\documentclass[11pt,twoside]{article}

\usepackage{epsfig,amsfonts,color}
\usepackage{amsmath}
\usepackage{mathtools}
\usepackage{amssymb, palatino, geometry,url}
\usepackage{amsthm}
\usepackage{threeparttable}
\usepackage{booktabs}
\usepackage{tabularx}

\newcommand{\bR}{\mathbb{R}}

\newcommand{\bU}{\mathcal{U}}
\newcommand{\bV}{\mathcal{V}}

\newcommand{\bL}{\mathcal{L}}

\newcommand{\rb}[1]{\left(#1\right)}
\newcommand{\bb}[1]{\left[#1\right]}

\usepackage[thinc]{esdiff}
\newcommand\norm[1]{\left\lVert#1\right\rVert}

\DeclarePairedDelimiter\floor{\lfloor}{\rfloor}

\usepackage{lineno}

\begin{document}

\title{{\color{black} Convolutional Neural Nets in Chemical Engineering:\\ Foundations, Computations, and Applications}}


\author{Shengli Jiang and Victor M. Zavala\thanks{Corresponding Author: victor.zavala@wisc.edu}\\
{\small Department of Chemical and Biological Engineering}\\
{\small \;University of Wisconsin-Madison, 1415 Engineering Dr, Madison, WI 53706, USA}}
 \date{}
\maketitle

\begin{abstract} {\color{black} In this paper we review the mathematical foundations of convolutional neural nets (CNNs) with the goals of}: i) highlighting connections with techniques from statistics, signal processing, linear algebra, differential equations, and optimization, ii) demystifying underlying computations, and iii) identifying new types of applications. CNNs are powerful machine learning models that highlight features from grid data to make predictions (regression and classification). The grid data object can be represented as vectors (in 1D), matrices (in 2D), or tensors (in 3D or higher dimensions) and can incorporate multiple channels (thus providing high flexibility in the input data representation). CNNs highlight features from the grid data by performing convolution operations with different types of operators. The operators highlight different types of features (e.g., patterns, gradients, geometrical features) and are learned by using optimization techniques. In other words, CNNs seek to identify optimal operators that best map the input data to the output data. A common misconception is that CNNs are only capable of processing image or video data but their application scope is much wider; specifically, datasets encountered in diverse applications can be expressed as grid data. Here, we show how to apply CNNs to new types of applications such as optimal control, flow cytometry, multivariate process monitoring, and molecular simulations. 
\end{abstract}


\section{Introduction}
Convolutional neural nets (CNNs) are powerful machine learning models that highlight (extract) features  from data to make predictions (regression and classification). The input data object processed by CNNs has a grid-like topology. {\color{black} For example, an image can be represented as a 2D grid data object that contains red, green, and blue (RBG) channels (each channel is a 2D matrix). Similarly, a video can be represented as a 3D grid data object (two spatial dimensions plus time) with RGB channels (each channel is a 3D tensor).} Features of the input object are extracted using a mathematical operation known as \textit{convolution}.  Convolutions are applied to the input data with different operators (also known as filters) that seek to extract different types of features (e.g., patterns, gradients, geometrical features). The goal of the CNN is to learn optimal operators (and associated features) that best map the input data to the output data. For instance, in recognizing an image (the input), the CNN seeks to learn the patterns of the image that best explains a label given to an image (the output).


The earliest version of a CNN was proposed in 1980 by Kunihiko Fukushima \cite{fukushima1980neocognitron} and was used for pattern recognition. In the late 1980s, the LeNet model proposed by LeCun et al. introduced the concept of \textit{backward propagation}, which streamlined learning computations using optimization techniques \cite{le1989handwritten}. Although the LeNet model had a simple architecture, it was capable of recognizing hand-written digits with high accuracy. In 1998, Rowley et al. proposed a CNN model capable of performing face recognition tasks  (this work revolutionized object classification and detection) \cite{rowley1998neural}. The complexity of CNN models (and their predictive power) has dramatically expanded with the advent of parallel computing architectures such as graphics processing units \cite{nickolls2008scalable}. Modern CNN models for image recognition include SuperVision \cite{krizhevsky2012imagenet}, GoogLeNet \cite{szegedy2014going}, VGG \cite{simonyan2014very}, and ResNet \cite{he2015deep}. New models are currently being developed to perform diverse computer vision tasks such as object detection \cite{ren2015faster}, semantic segmentation \cite{long2015fully}, action recognition \cite{simonyan2014two}, and 3D analysis \cite{ji20123d}. Nowadays, CNNs are routinely used in applications such as in the face-unlock feature of smartphones \cite{network2017face}. 

CNNs tend to outperform other machine learning models (e.g., support vector machine and decision tree) but their behavior is difficult to explain. For instance, it is not always straightforward to determine the features that the operators are seeking to extract. As such, researchers have devoted significant effort into understanding the mathematical properties of CNNs. For instance, Cohen et al. established an equivalence between CNNs and hierarchical tensor factorizations \cite{cohen2016expressive}. Bruna et al. analyzed feature extraction capabilities \cite{bruna2013invariant} and showed that these satisfy translation invariance and deformation stability (important concepts in determining geometric features from data). 

While CNNs were originally developed to perform computer vision tasks, the grid data representation used by CNNs is flexible and can be used to process datasets arising in many different applications. For instance, in the field of chemistry, Hirohara et al. proposed a matrix representations of SMILES strings (which encodes molecular topology) by using a technique known as one-hot encoding \cite{hirohara2018convolutional}. The authors used this representation to train a CNN that could predict  the toxicity of chemicals; it was shown that the CNN outperformed traditional models based on  fingerprints (an alternative molecular representation). Via analysis of the learned filters, the authors also  determined chemical structures (features) that drive toxicity. In the realm of biology, Xie et al. applied CNNs to count and detect cells from micrographs \cite{xie2018microscopy}.  In area of material science, Smith et al. have used CNNs to extract features from optical micrographs of liquid crystals to design chemical sensors \cite{smith2020convolutional}.

While there has been extensive research on CNNs and the number of applications in science and engineering is rapidly growing, there are limited reports available in the literature that outline the mathematical foundations and operations behind CNNs. As such, important connections between CNNs and other fields such as statistics, signal processing, linear algebra, differential equations, and optimization remain under-appreciated. We believe that this disconnect limits the ability of researchers to propose extensions and identify new applications. For example, a common misconception is that CNNs are only applicable to computer vision tasks; however, CNNs can operate on general grid data objects (vectors, matrices, and tensors). Moreover, operators learned by CNNs can be potentially explained when analyzed from the perspective of calculus and statistics. For instance, certain types of operators seek to extract gradients (derivatives) of a field or seek to extract correlation structures. Establishing connections with other mathematical fields is important in gaining interpretability of CNN models. Understanding the operations that take place in a CNN is also important in order to understand their inherent limitations, in proposing potential modeling and algorithmic extensions, and in facilitating the incorporation of CNNs in computational workflows of interest to engineers (e.g., process monitoring, control, and optimization). 

In this work, we review the mathematical foundations of CNNs; specifically, we provide concise derivations of input data representations and of convolution operations. We explain the origins of certain types of operators and the data transformations that they induce to highlight features that are hidden in the data. Moreover, we provide concise derivations for forward and backward propagations that arise in CNN training procedures. These derivations provide insight into how information flows in the CNN and help understand computations involved in the learning process (which seeks to solve an optimization problem that minimizes a loss function). We also explain how derivatives of the loss function can be used to understand key features that the CNN searches for to make predictions.  We illustrate the concepts by applying CNNs to new types of applications such as optimal control (1D) , flow cytometry (2D), multivariate process monitoring (2D), and molecular simulations (3D).  Specifically, we focus our attention on how to convert raw data into a suitable grid-like representation that the CNN can process. 

{\color{black} The paper is structured as follows. In Section \ref{sec:conv_operation} we describe the mathematical foundations of convolution operations. In Section \ref{sec:conv_operator}  we discuss convolution operations used in CNNs. In Section \ref{sec:cnn} we describe the building blocks of a typical CNN architecture. In Section \ref{sec:cnn_training} we describe the training process of a CNN, including the forward and backward propagation. Section \ref{sec:cases} presents applications of CNNs  to chemical engineering problems. Section \ref{sec:conclusion} presents conclusions and suggests directions of future work.}

\section{Convolution Operations}\label{sec:conv_operation}

Convolution is a mathematical operation that involves an input function and an operator function (these functions are also known as {\it signals}). A related operation that is often used in CNNs is cross-correlation. Although technically different, the essence of these operations is similar (we will see that they are mirror representations). One can think of a convolution as an operation that seeks to transform the input function in order to highlight (or extract) features. The input and operator functions can live on arbitrary dimensions (1D, 2D, 3D, and higher). The features highlighted by an operator are defined by its design; for instance, we will see that one can design operators that extract specific patterns, derivatives (gradients), or frequencies. These features encode different aspects of the data (e.g., correlations and geometrical patterns) and are often hidden. Input and operator functions can be continuous or discrete; discrete functions facilitate computational implementation but continuous functions can help understand and derive mathematical properties.  For example, families of discrete operators can be generated by using a single continuous function (a kernel function).  We begin our discussion with 1D convolution operations and we later extend the analysis to the 2D case; extensions to higher dimensions are straightforward once the basic concepts are outlined.  We highlight, however, that convolutions in high dimensions are computationally expensive (and sometimes intractable). 

\subsection{1D Convolution Operations}
The convolution of scalar continuous functions $u: \bR \mapsto \bR$ and $v: \bR \mapsto \bR$ is denoted as $\psi = u * v$ and their cross-correlation is denoted as $\phi = u \star v$. These operations are given by:
\begin{subequations}
\begin{align}
	\psi \rb{x} & = (u*v)(x)= \int_{-\infty}^{\infty} u\rb{x'}\cdot v\rb{x-x'}dx', \; x \in (-\infty, \infty)      \\
	\phi \rb{x} & = (u\star v)(x)=\int_{-\infty}^{\infty} u\rb{x'}\cdot v\rb{x+x'}dx', \; x \in (-\infty, \infty).
\end{align}
\end{subequations}
We will refer to function $u$ as the convolution operator and to $v$ as the input signal. The output of the convolution operation is a scalar continuous function $\psi: \bR \mapsto \bR$ that we refer to as the {\em convolved signal}. The output of the cross-correlation operation is also a continuous function $\phi: \bR \mapsto \bR$ that we refer to as the {\em cross-correlated signal}.    
\\

The convolution and cross-correlation are applied to the signal by spanning the domain $x\in (-\infty,\infty)$; we can see that convolution is applied  by looking {\it  backward}, while cross-correlation is applied by looking {\it  forward}.  One can show that cross-correlation is equivalent to convolution that uses a rotated (flipped) operator $u$ by 180 degrees; in other words, if we define the rotated operator as $-u$ then $u*v=(-u)\star v$. We thus have that convolution and cross-correlation are mirror operations; consequently, convolution and cross-correlation are often both referred to as {\it  convolution}.   In what follows we use the term convolution and cross-correlation interchangeably; we highlight one operation over the other when appropriate.  Modern CNN packages such as PyTorch \cite{NEURIPS2019_9015} use cross-correlation in their implementation (this is more intuitive and easier to implement). 
\\

\begin{figure}[!htb]
\begin{center}
\includegraphics[width=0.6\linewidth]{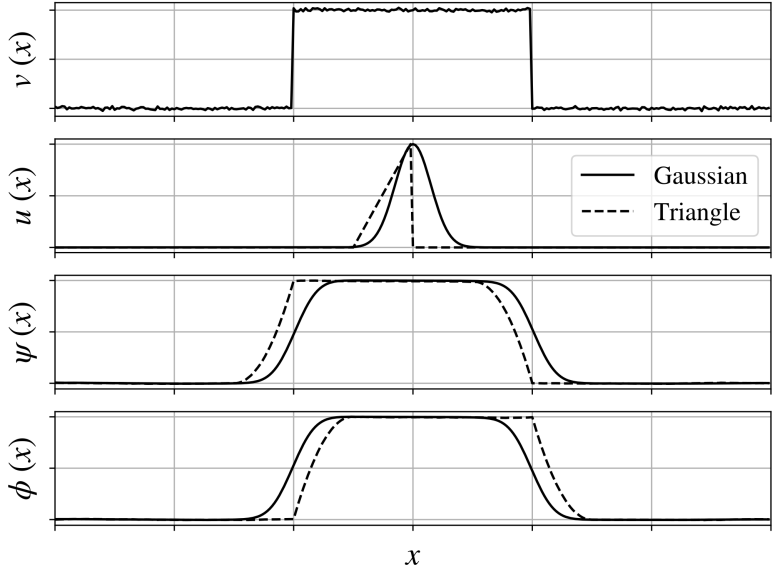}
\caption{From top to bottom: Input signal $v$, Gaussian and triangular operators $u$,  convolved signal $\psi$, and cross-correlated signal $\phi$. }
\label{fig:1d_conv_corr}
\end{center}
\end{figure}

One can think of convolution and cross-correlation as {\it  weighting} operations (with weighting function defined by the operator).  The weighting function $u$ can be designed in different ways to perform different operations to the signal $v$ such as averaging, smoothing, differentiation, and pattern recognition. In Figure \ref{fig:1d_conv_corr} we illustrate the application of a Gaussian operator $u$ to a noisy rectangular signal $v$. We note that the convolved and cross-correlated signals are identical (because the operator is symmetric and thus $u=-u$). We also note that the output signals are smooth versions of the input signal; this is because a Gaussian operator has the effect of extracting frequencies from the signal (a fact that can be established from Fourier analysis).  Specifically, we recall that the Fourier transform of the convolved signal $\psi$ satisfies:
\begin{equation}
\mathcal{F}\{\psi\}=\mathcal{F}\{u*v\}=\mathcal{F}\{u\}\cdot \mathcal{F}\{v\}.
\end{equation}
where $\mathcal{F}\{w\}$ is the Fourier transform of scalar function $w$. Here, the product $\mathcal{F}\{u\}\cdot \mathcal{F}\{v\}$ has the effect of preserving the frequencies in the signal $v$ that are also present in the operator $u$. In other words, convolution acts as a filter of frequencies; as such, one often refers to operators as filters. By comparing the output signals of convolution and cross-correlation obtained with the triangular operator, we can confirm that one is the mirror version of the other. From Figure \ref{fig:1d_conv_corr}, we also see that the output signals are smooth variants of the input signal; this is because the triangular operator also extracts frequencies from the input signal (but the extraction is not as clean as that obtained with the Gaussian operator). This highlights the fact that different operators have different frequency content (known as the frequency spectrum). 
\\

Convolution and cross-correlation are implemented computationally by using discrete representations;  these signals are represented by column vectors $u \in \bR^{n}$ and $v \in \bR^{n}$ with entries denoted as  $u[x]$, $v[x]$, $x=-N,...,N$ (note that $n=2\cdot N+1$). Here, we note that the input and operator are defined over a 1D grid. The discrete convolution results in vector $\psi\in \bR^n$ with entries given by:
\begin{equation}
\psi \bb{x} = \sum_{x'=-N}^{N} u\bb{x'}\cdot v\bb{x+x'}, \; x \in \{-N,N\}.
\end{equation}
Here, we use $\{-N,N\}$ to denote the integer sequence $-N,-N+1,...,-1,0,1,...,N-1,N$.  We note that computing $\psi \bb{x} $ at a given location $x$ requires $n$ operations and thus computing the entire output signal (spanning $x\in \{-N,N\}$) requires $n^2$ operations;  consequently, convolution becomes expensive when the signal and operator are long vectors; as such, the operator $u$ is often a vector of low dimension (compared to the dimension of the signal $v$). Specifically, consider the operator $u \in \bR^{n_u}$ with entries $u[x],\,x=-N_u,...,N_u$ (thus $n_u=2\cdot N_u+1$) and assume $n_u\ll n$ (typically $n_u=3$). The convolution is given by:
\begin{equation}
\psi \bb{x} = \sum_{x'=-N_u}^{N_u} u\bb{x'}\cdot v\bb{x+x'}, \; x \in \{-N,N\}.
\end{equation}
This reduces the number of operations needed from $n^2$ to $n\cdot n_u$.  
\\

\begin{figure}[!htb]
	\begin{center}
		\includegraphics[width=0.8\linewidth]{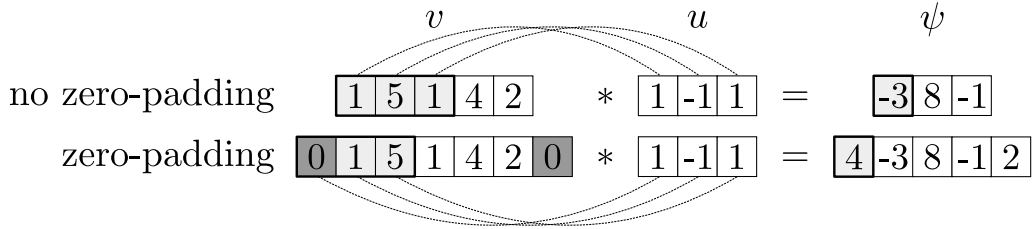}
		\caption{{\color{black} Illustration of 1D convolution with (bottom) and without (top) zero-padding.}}
		\label{fig:zeropadding}
	\end{center}
\end{figure}

The convolved signal is obtained by using a {\it  moving window} starting at the boundary $x=-N$ and by moving forward as $x+1$ until reaching $x=N$. We note that the convolution is not properly defined close to the boundaries (because the window lies outside the domain of the signal). This situation can be remedied by starting the convolution at an inner entry of $v$ such that the full window is within the signal (this gives a {\it  proper} convolution). However, this approach has the effect of returning a signal $\psi$ that is smaller than the original signal $v$. This issue can be overcome by artificially {\it  padding} the signal by adding zero entries (i.e., adding ghost entries to the signal at the boundaries). This is an {\it  improper} convolution but returns a signal $\psi$ that has the same dimension as $v$ (this is often convenient for analysis and computation).  Figure \ref{fig:zeropadding} illustrates the difference between convolutions with and without padding. We also highlight that it is possible to advance the moving window by multiple entries as $x+s$ (here, $s\in \mathbb{Z}_+$ is known as the {\it  stride}). This has the effect of reducing the number of operations (by skipping some entries in the signal) but returns a signal that is smaller than the original one.

By defining signals $u\in \mathbb{R}^{n_u}$ and $v\in \mathbb{R}^{n_v}$ with entries $u[x], \ x=1,...,n_u$ and $v[x], \ x=1,...,n_v$, one can also express the valid convolution operation in an {\it  asymmetric} form as:
\begin{equation}\label{eq:siso}
\psi \bb{x} = \sum_{x'=1}^{m} u\bb{x'}\cdot v\bb{x+x'-1}, \; x \in \{1, n_v-n_u+1\}. 
\end{equation}
\\

In CNNs, one often applies multiple operators to a single input signal or one analyzes input signals that have multiple channels. This gives rise to the concepts of single-input single-output (SISO), single-input multi-output (SIMO), and multi-input multi-output (MIMO) convolutions. 
\\

In SISO convolution, we take a 1-channel input vector and output a vector (a 1-channel signal), as in \eqref{eq:siso}. In SIMO convolution, one uses a single input $v\in\mathbb{R}^{n_v}$ and a collection of $q$ operators $\mathcal{U}_{\rb{j}} \in \bR^{n_u}$ with $j\in \{1,q\}$. Here, every element $\mathcal{U}_{\rb{j}}$ is an $n_u$-dimensional vector and we represent the collection as the object $\mathcal{U}$. SIMO convolution yields a collection of convolved signals $\Psi_{\rb{j}} \in \bR^{n_v-n_u+1}$ with $j\in \{1,q\}$ and with entries given by:
\begin{equation}
\Psi_{\rb{j}} \bb{x} = \sum_{x'=1}^{m} \mathcal{U}_{\rb{j}}\bb{x'}\cdot v\bb{x+x'-1}, \; x \in \{1, n_v-n_u+1\}.  
\end{equation}

In MIMO convolution, we consider a multi-channel input given by the collection $\mathcal{V}_{\rb{i}}\in \mathbb{R}^{n_v},\,i\in \{1,p\}$ ($p$ is the number of channels); the collection is represented as the object $\mathcal{V}$. We also consider a collection of operators $\bU_{(i,j)}\in \bR^{n_u}$ with $i\in \{1,p\}$ and $j\in \{1,q\} $; in other words, we have $q$ operators per input channel $i\in \{1,p\}$ and  we represent the collection as the object $\mathcal{U}$. MIMO convolution yields the collection of convolved signals $\Psi_{\rb{j}} \in \bR^{n_v-n_u+1}$ with $j\in \{1,q\}$ and with entries given by:
\begin{equation}
\begin{aligned}
	\Psi_{\rb{j}} \bb{x} & = \sum_{i=1}^{p} \bU_{(i,j)} *\mathcal{V}_{\rb{i}}                                                                      \\
	                     & = \sum_{i=1}^{p} \sum_{x'=1}^{m} \bU_{(i,j)}\bb{x'} \cdot \mathcal{V}_{\rb{i}} \bb{x+x'-1}, \; x \in 
    \{1, n_v-n_u+1\}.
\end{aligned}
\end{equation}
We see that, in MIMO convolution, we add the contribution of all channels $i \in \{1,p\}$ to obtain the output object $\Psi$, which contains $j \in \{1,q\}$ channels given by vectors $\Psi_{(i)}$ of dimension $(n_v-n_u+1)$. Channel combination loses information but saves computer memory.

\subsection{2D Convolution Operations}\label{sec:2d_conv_and_corr}

Convolution and cross-correlation operations in 2D are analogous to those in 1D; the convolution and cross-correlation of a continuous operator $u: \bR^2 \mapsto \bR$ and an input signal $v: \bR^2 \mapsto \bR$ are denoted as $\psi = u * v$ and $\phi=u\star v$ and are given by:
\begin{subequations}
\begin{align}
	\psi \rb{x_1, x_2} & = \int_{-\infty}^{\infty} \int_{-\infty}^{\infty} u\rb{x_1', x_2'}\cdot v\rb{x_1-x_1', x_2-x_2'}dx_1'dx_2', \qquad x_1, x_2 \in \rb{-\infty, \infty} \\
	\phi \rb{x_1, x_2} & = \int_{-\infty}^{\infty} \int_{-\infty}^{\infty} u\rb{x_1', x_2'}\cdot v\rb{x_1+x_1', x_2+x_2'}dx_1'dx_2', \qquad x_1, x_2 \in \rb{-\infty, \infty}
\end{align}
\end{subequations}
As in the 1D case, the terms convolution and cross-correlation are used interchangeably; here, we will use the cross-correlation form (typically used in CNNs). 
\\

\begin{figure}[!htb]
	\begin{center}
		\includegraphics[width=0.8\linewidth]{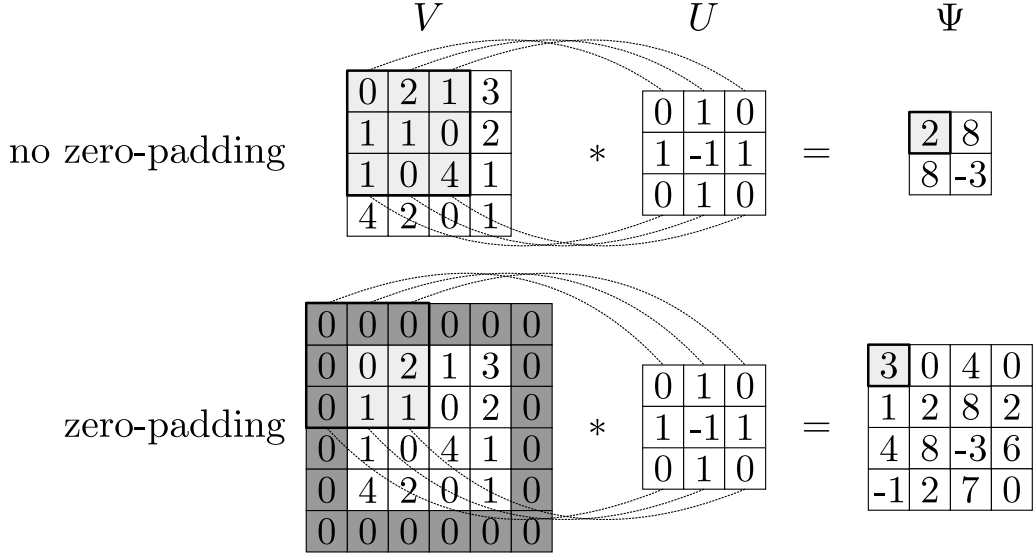}
		\caption{{\color{black} Illustration of 2D convolution with (bottom) and without (top) zero padding.}}
		\label{fig:2dzeropadding}
	\end{center}
\end{figure}

In the discrete case, the input signal and the convolution operator are {\it  matrices} $U \in \bR^{n_U\times n_U}$ and $V\in \bR^{n_V\times n_V}$ with entries $U\bb{x_1,x_2}, \; x_1,x_2 \in \{-N_U, N_U\}$ and $V\bb{x_1,x_2}, \; x_1,x_2 \in \{-N_V,N_V\}$ (thus $n_U=2\cdot N_U+1$ and $n_V=2\cdot N_V+1$). For simplicity, we assume that these are square matrices and we note that the input and operator are defined over a 2D grid. The convolution of the input and operator matrices results in a matrix $\Psi=U*V$ with entries: 
\begin{equation}
\Psi \bb{x_1,x_2} = \sum_{x'_1=-N_U}^{N_U} \sum_{x'_2=-N_U}^{N_U}U\bb{x_1',x_2'}\cdot V\bb{x_1+x_1',x_2+x_2'}, \; x_1,x_2 \in \{-N_V,N_V\}.
\end{equation}
In Figure \ref{fig:2dzeropadding} we illustrate 2D convolutions with and without padding. The 2D convolution (in valid and asymmetric form) can be expressed as:
\begin{align}\label{eq:2dSISO}
	\Psi \bb{x_1, x_2} &= \sum_{x_1'=1}^{n_U}\sum_{x_2'=1}^{n_U} U\bb{x_1', x_2'}\cdot V\bb{x_1+x_1'-1, x_2+x_2'-1},
\end{align}
where $x_1 \in \{1, n_V-n_U+1\}$, $x_2 \in \{1, n_V-n_U+1\}$ and $\Psi \in \bR^{\rb{n_V-n_U+1} \times \rb{n_V-n_U+1}}$.

The SISO convolution of an input $V \in \bR^{n_V \times n_V}$ and an operator $U \in \bR^{n_U \times n_U}$ is given by \eqref{eq:2dSISO} and outputs a matrix $\Psi \in \bR^{\rb{n_V-n_U+1} \times \rb{n_V-n_U+1}}$.  In SIMO convolution, we are given a collection of operators $\bU_{\rb{j}} \in \bR^{n_\mathcal{U} \times n_\mathcal{U}}$ with $j\in \{1,q\}$. A convolved matrix $\Psi_{\rb{j}} \in \bR^{\rb{n_{V}-n_\mathcal{U}+1} \times \rb{n_{V}-n_\mathcal{U}+1}}$ is obtained by applying the $j$-th operator $\bU_{\rb{j}}$ to the input $V$:
\begin{align}
\Psi_{\rb{j}} \bb{x_1, x_2} = \sum_{x_1' = 1}^{n_\mathcal{U}} \sum_{x_2' = 1}^{n_\mathcal{U}} \bU_{\rb{j}} \bb{x_1', x_2'} V\bb{x_1+x_1'-1, x_2+x_2'-1} 
\end{align}
for $j\in\{1, q\}, \; x_1,x_2 \in \{1, n_{V}-n_\mathcal{U}+1\}$.  The collection of convolved matrices $\Psi_{\rb{j}}\in \bR^{\rb{n_{V}-n_\mathcal{U}+1} \times \rb{n_{V}-n_\mathcal{U}+1}},\; \ j\in\{1,q\}$ is represented as object  $\Psi$. 
\\

In MIMO convolution, we are given a $p$-channel input collection $\bV_{\rb{i}}\in \bR^{n_\mathcal{V} \times n_\mathcal{V}}$ with $i\in \{1,p\}$ and represented as $\bV$. We convolve this input with the operator object $\bU$, which is a collection $\bU_{(i,j)} \in \bR^{n_\mathcal{U} \times n_\mathcal{U} },\;i\in\{1,p\},\; j\in \{1,q\}$. This results in an object $\Psi$ given by the collection $\Psi_{(j)} \in \bR^{\rb{n_\mathcal{V}-n_\mathcal{U}+1} \times \rb{n_\mathcal{V}-n_\mathcal{U}+1}}$ for $j\in \{1,q\}$ and with entries given by:
\begin{equation}
\begin{aligned}
	\Psi_{\rb{j}} \bb{x_1, x_2} 
	& =\sum_{i=1}^{p}\bU_{(i,j)}*\bV_{\rb{i}}\\
	& = \sum_{i=1}^{p}\sum_{x_1' = 1}^{n_\mathcal{U}} \sum_{x_2' = 1}^{n_\mathcal{U}} \bU_{(i,j)}\bb{x_1', x_2'} \bV_{\rb{i}}\bb{x_1+x_1'-1, x_2+x_2'-1},
\end{aligned}
\end{equation}
for $j\in\{1,q\},\; x_1\in \{1,n_\mathcal{V}-n_\mathcal{U}+1\}$, and $x_2\in \{1,n_\mathcal{V}-n_\mathcal{U}+1\}$.  For convenience, the input object is often represented as a 3D tensor $\bV\in \bR^{n_\mathcal{V} \times n_\mathcal{V}\times p}$, the convolution operator is represented as the 4D tensor $\bU\in \bR^{n_\mathcal{U} \times n_\mathcal{U}\times p \times q}$, and the output signal is represented as the 3D tensor $\Psi \in \bR^{\rb{n_\mathcal{V}-n_\mathcal{U}+1} \times \rb{n_\mathcal{V}-n_\mathcal{U}+1}\times q}$. We note that, if $p=1$ and $q=1$ (1-channel inputs and channels), these tensors become matrices. Tensors are high-dimensional quantities that require significant computer memory to store and significant power to process. 

\begin{figure}[!htb]
	\begin{center}
		\includegraphics[width=0.8\linewidth]{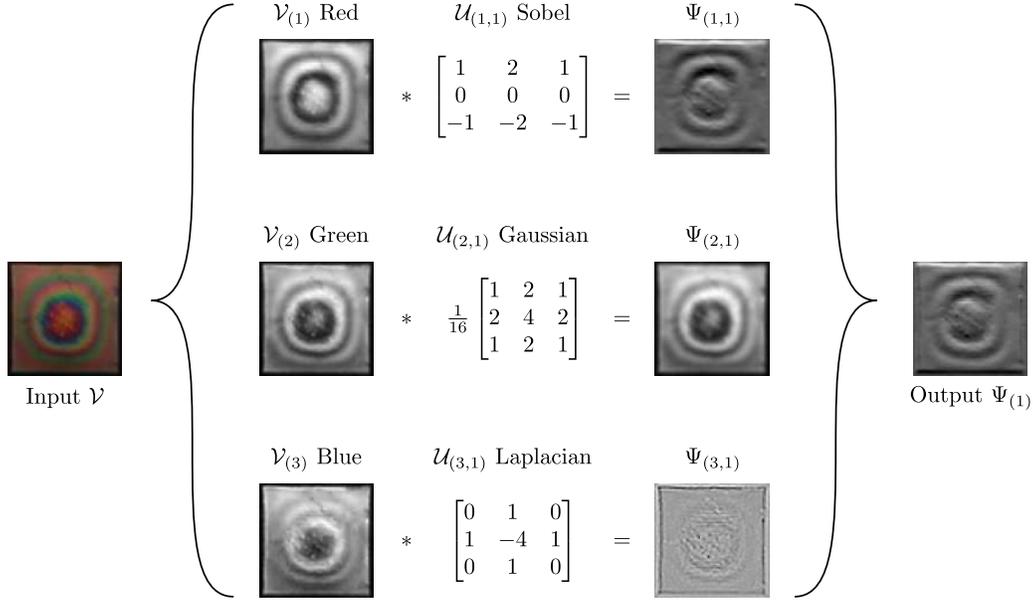}
		\caption{MIMO convolution of an RGB image $\mathcal{V} \in \bR^{50 \times 50 \times 3}$ (3-channel input) corresponding to a liquid crystal micrograph. The red, green, and blue (RGB) channels of the input 3D tensor $\mathcal{V}$ are convolved with an operator $\mathcal{U} \in \bR^{3 \times 3 \times 3 \times 1}$  (a 3D tensor since $q=1$) that contains a Sobel operator (for edge detection), Gaussian operator (for blurring), and Laplacian operator (for edge detection). The output $\Psi \in \bR^{48 \times 48 \times 1}$ is a matrix (since $q=1$) that results from the combination of the convolved matrices.}
		\label{fig:rgbimages}
	\end{center}
\end{figure}
\vspace{0.1in}

MIMO convolutions in 2D are typically used to process RGB images (3-channel input), as shown in Figure \ref{fig:rgbimages}. Here, the RGB image is the object $\bV$ and each input channel $\bV_{\rb{i}}$ is a matrix; each of these matrices is convolved with an operator $\bU_{\rb{i,1}}$. Here we assume $q=1$ (one operator per channel). The collection of convolved matrices  $\Psi_{\rb{i,1}}$ are combined to obtain a single matrix $\Psi_{\rb{1}}$.  If we consider a collection of operators $\bU_{\rb{i,j}}$ with $i\in\{1,p\}$, $j\in\{1,q\}$ and $q>1$,  the output of MIMO convolution returns the collection of matrices $\Psi_{\rb{j}}$ with $j\in\{1,q\}$, which is assembled in the tensor $\Psi$. Convolution with multiple operators allows for the extraction of different types of features from different channels.
\\

 We highlight that the use of channels is not restricted to images; specifically, channels can be used to input data of different variables in a grid  (e.g., temperature, concentration, density). As such, channels provide a flexible framework to express multivariate data.  
 \\
 
 We also highlight that a grayscale image is a 1-channel input matrix (the RGB channels are combined in a single channel). In a grayscale image, every pixel (an entry in the associated matrix) has a certain light intensity value; whiter pixels have higher intensity and darker pixels have a lower intensity (or the other way around). The resolution of the image is dictated by the number of pixels and thus dictates the size of the matrix; the size of the matrix dictates the amount of memory needed for storage and computations needed for convolution.  It is also important to emphasize that {\it  any matrix can be visualized as a grayscale image} (and any grayscale image has an underlying matrix). This {\it  duality} is important because visualizing large matrices (e.g., to identify patterns) is difficult if one directly inspects the numerical values; as such, a convenient approach to analyze patterns in large matrices consists of visualizing them as images. 
\\

Finally, we highlight that convolution operators can be applied to any grid data object in higher dimensions (e.g., 3D and 4D) in a similar manner. Convolutions in 3D can be used to process video data (each time frame is an image). However, 3D data objects can also be used to represent data distributed over 3D Euclidean spaces (e.g., density or flow in a 3D domain). However, the complexity of convolution operations in 3D (and higher dimensions) is substantial.  In the discussion that follows we focus our attention to 2D convolutions; in Section \ref{sec:cases} we illustrate the use of 3D convolutions in a practical application. 

\section{Convolution Operators}\label{sec:conv_operator}

Convolution operators are the key functional units that a CNN uses to extract features from input data objects. Some commonly used operators and their transformation effect are shown in Figure \ref{fig:rgbimages}. When the input is convolved with a Sobel operator, the output highlights the edges (gradients of intensity). The reason for this is that the Sobel operator is a {\it  differential operator}. To explain this concept, consider a discrete 1D signal $v$; its derivative at entry $x$ can be computed using the finite difference:
\begin{equation}
	v'[x] = \frac{v[x+1]-v[x-1]}{2}.
\end{equation}
This indicates that we can compute the derivative signal $v'$ by applying a convolution of the form $v'=u*v$ with operator $u=(1/2,0,-1/2)$. Here, the operator can be scaled as $u=(1,0,-1)$ as this does not alter the nature of the operation performed (just changes the scale of the entries of $v'$). 
\\

The Sobel operator shown in Figure~\ref{fig:rgbimages} is a $3\times 3$ matrix of the form: 
\begin{equation}
	U=\begin{bmatrix}
		1  & 2  & 1  \\
		0  & 0  & 0  \\
		-1 & -2 & -1
	\end{bmatrix}
	=
	\begin{bmatrix}
		1  \\
		0  \\
		-1
	\end{bmatrix}
	\begin{bmatrix}
		1 & 2 & 1
	\end{bmatrix},
\end{equation}
where the vector $(1, 0, -1)$ approximates the first derivative in the vertical direction, and $(1, 2, 1)$ is a binomial operator that smooths the input matrix.  
\\

Another operator commonly used to detect edges in images is the Laplacian operator; this operator is an approximation of the continuous Laplacian operator $\mathcal{L} =\frac{\partial^2 }{\partial x_1^2}+\frac{\partial ^2}{\partial x_2^2}$. The convolution of a matrix $V$ using a $3\times 3$ Laplacian operator $U$ is:
\begin{equation}
	\begin{aligned}
		U*V & = V\bb{x_1-1, x_2} + V\bb{x_1+1, x_2} + V\bb{x_1, x_2-1}+V\bb{x_1, x_2+1}-4\cdot V\bb{x_1, x_2}.
	\end{aligned}
\end{equation}
This reveals that the convolution is an approximation of $\mathcal{L} $ that uses a 2D finite difference scheme; this scheme has an operator of the form: 
\begin{equation}
	U=\begin{bmatrix}
		0 & 1  & 0 \\
		1 & -4 & 1 \\
		0 & 1  & 0
	\end{bmatrix}.
\end{equation}
The transformation effect of the Laplacian operator is shown in Figure~\ref{fig:rgbimages}.  In the partial differential equations (PDE) literature, the non-zero structure of a finite-difference operator is known as the {\it  stencil} \cite{sauer2018numerical}. As expected, a wide range of finite difference approximations (and corresponding operators) can be envisioned.  Importantly, since the Laplacian operator computes the second derivative of the input, this is suitable to detect locations of minimum or maximum intensity in an image (e.g., peaks).  This allows us to understand the geometry of matrices (e.g., geometry of images or 2D fields). Moreover, this also reveals connections between PDEs and convolution operations. 
\\

The Gaussian operator is commonly used to smooth out images. A Gaussian operator is defined by the  density function: 
\begin{equation}
	U\rb{x_1, x_2} = \frac{1}{2 \pi \sigma^2} e^{-\frac{x_1^2 + x_2^2}{2\sigma^2}},
\end{equation}
where $\sigma\in \mathbb{R}_+$ is the standard deviation. The standard deviation determines the spread of of the operator and is used to control the frequencies removed from a signal. We recall that the density of a Gaussian is maximum at the center point and decays exponentially (and symmetrically) when moving away from the center. Discrete representations of the Gaussian operator are obtained by manipulating $\sigma$ and truncating the density in a window. For instance, the $3\times 3$ Gaussian operator as shown in Figure~\ref{fig:rgbimages} is:
\begin{equation}
	U=\frac{1}{16}
	\begin{bmatrix}
		1 & 2 & 1 \\
		2 & 4 & 2 \\
		1 & 2 & 1
	\end{bmatrix}.
\end{equation}
Here, we can see that the operator is square and symmetric, has a maximum value at the center point, and the values decay rapidly as one moves away from the center.  
\\

\begin{figure}[!htb]
	\begin{center}
		\includegraphics[width=0.6\linewidth]{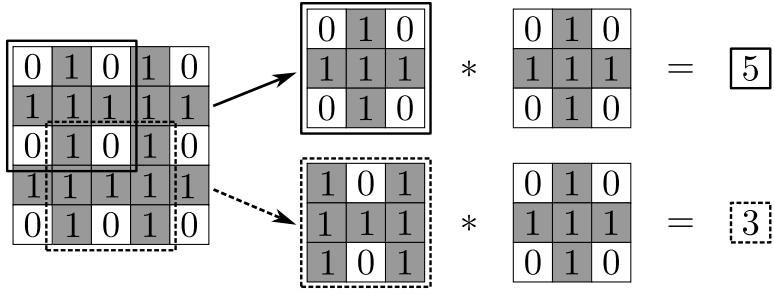}
		\caption{Highlighting sought patterns by convolutional operators. When the sought pattern is matched, the convolution generates a large signal.}
		\label{fig:patdetect}
	\end{center}
\end{figure}

Convolution operators can also be designed to perform pattern recognition; for instance, consider the situation in which you want to highlight areas in a matrix that have a pattern (feature) of interest. Here, the structure of the operator dictates the 0-1 pattern sought. For instance, in Figure~\ref{fig:patdetect} we present an input matrix with 0-1 entries and we perform a convolution with an operator with a pre-defined 0-1 structure. The convolution highlights the areas in the input matrix in which the presence of the sought pattern is strongest (or weakest). As one can imagine, a huge number of operators could be designed to extract different patterns (including smoothing and differentiation operators); moreover, in many cases it is not obvious which patterns or features might be present in the image. This indicates that one requires a systematic approach to automatically determine which features of an input signal (and associated operators) are most relevant.
\\

\section{CNN Architectures}\label{sec:cnn}
CNNs are hierarchical (layered) models that perform a sequence of convolution, activation, pooling, flattening operations to extract features from input data object. The output of this sequence of {\it  transformation} operations is a vector that summarizes the feature information of the input; this feature vector is fed to a fully-connected neural net that makes final predictions. The goal of the CNN is to determine the features of a set of input data objects (input data samples) that best predict the corresponding output samples. Specifically, the input of the CNN are a set of sample objects $\bV^{(i)}\in \bR^{n_\mathcal{V} \times n_\mathcal{V} \times p}$ with sample index $i\in\{1,n\}$, and the output of the CNN are the corresponding predicted labels $\hat{y}[i],\; i\in\{1,n\}$. The goal of a CNN is to determine convolution operators giving features that best match the predicted labels to the output labels; in other words, the CNN seeks to find operators (and associated features) that best map the inputs to the outputs. 
\\

\begin{figure}[!htb]
	\begin{center}
		\includegraphics[width=1\linewidth]{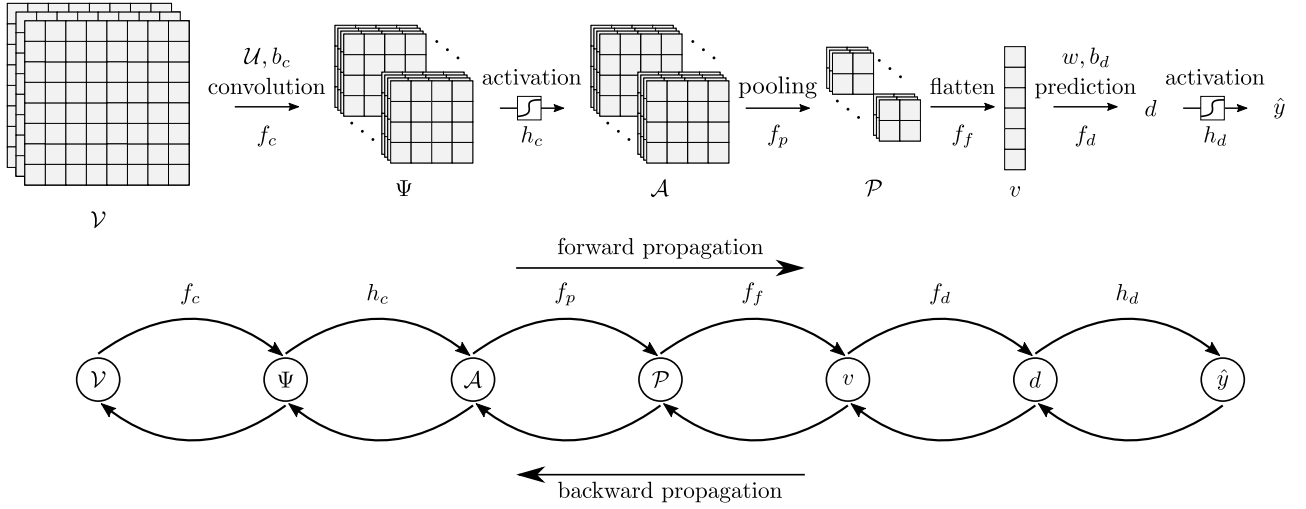}
		\caption{CNN architecture (top) comprised of a convolution block $f_c$, a max-pooling block $f_{p}$, a flattening block  $f_f$, and a dense block $f_d$. Here, $h_c$ and $h_d$ are activations applied in the convolution and dense blocks, respectively. A sample input is a $p$-channel object $\bV \in \bR^{n_\mathcal{V} \times n_\mathcal{V} \times p}$ and the predicted output is a scalar $\hat{y} \in \bR$. The parameters of the CNN are $\bU$, $b_c$ (for convolution block) and $w$, $b_d$ (for dense block). During training (bottom), forward propagation processes the sample input and outputs a scalar, and backward propagation calculates the gradients by recursively using chain-rules from the output to the input.}
		\label{fig:architecture}
	\end{center}
\end{figure}

In this section we discuss the different elements of a 2D CNN architecture. To facilitate the analysis, we construct a simple CNN of the form shown in Figure \ref{fig:architecture}. This architecture contains a single layer that performs convolution, activation, pooling, and flattening. Generalizing the discussion to multiple layers and higher dimensions (e.g., 3D) is rather straightforward once the basic concepts are established. Also, in the discussion that follows, we consider a single input data sample (that we denote as $\mathcal{V}$) and discuss the different transformation operations performed to it along the CNN to obtain a final prediction (that we denote as $\hat{y}$). We then discuss how to combine multiple samples to train the CNN. {\color{black} We emphasize that our goal here is to focus on the foundations of CNNs; discussing more complex architectures (e.g., with more layers or multiple outputs) requires introducing additional notation and concepts that we believe will blur our main message}.  

\subsection{Convolution Block}

\begin{figure}[!htb]
	\begin{center}
		\includegraphics[width=0.8\linewidth]{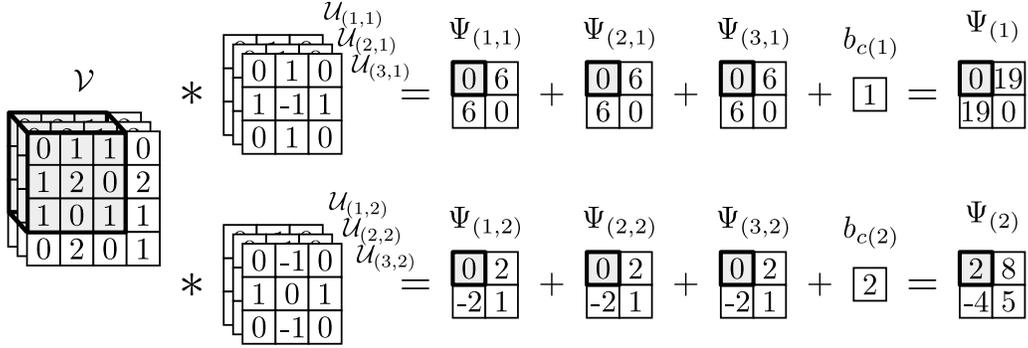}
		\caption{Convolution block for a 3-channel input $\bV \in \bR^{4 \times 4 \times 3}$ and a 2-channel operator $\bU \in \bR^{3\times 3 \times 3 \times 2}$. The output $\Psi \in \bR^{2 \times 2 \times 2}$ is obtained by combining the contributions of the different channels. A bias $b_c$ is added to this combination to give the final output of the convolution block.}
		\label{fig:conv_layer}
	\end{center}
\end{figure}

An input sample of the CNN is the tensor $\bV \in \bR^{n_\mathcal{V} \times n_\mathcal{V} \times p}$. The convolution block  uses the operator $\bU \in \bR^{n_\mathcal{U} \times n_\mathcal{U} \times p \times q}$ to conduct a MIMO convolution. The output of this convolution is the tensor $\Psi\in \bR^{n_\Psi \times n_\Psi\times q}$ with entries given by:
\begin{equation}
\begin{aligned}
	\Psi_{\rb{j}} \bb{x_1, x_2} & = b_{c}[j] + \sum_{i=1}^{p} \sum_{x_1' = 1}^{n_\mathcal{U}} \sum_{x_2' = 1}^{n_\mathcal{U}} \bU_{(i,j)}\bb{x_1', x_2'} \bV_{\rb{i}}\bb{x_1+x_1'-1, x_2+x_2'-1},
\end{aligned}
\end{equation}
where $x_1 \in \{1, n_\Psi\}$, $x_2 \in \{1, n_\Psi\}$, and $j\in \{1,q\}$. A bias parameter $b_{c}[j] \in \bR$ is added after the convolution operation; the bias helps adjust the magnitude of the convolved signal.  To enable compact notation, we define the convolution block using the mapping $\Psi= f_{c}(\bV; \bU,b_c)$; here, we note that the mapping depends on the parameters $\mathcal{U}$ (the operators) and $b_c$ (the bias). An example of a convolution block with a 3-channel input and 2-channel operator is shown in Figure~\ref{fig:conv_layer}. 
\\

\subsection{Activation Block}

The convolution block outputs the signal $\Psi$; this is passed through an activation block given by the mapping $\mathcal{A} = h_c(\Psi)$ with $\mathcal{A}\in  \bR^{n_\Psi \times n_\Psi \times q}$. Here, we define an activation mapping of the form $h_c: \bR^{n_\Psi \times n_\Psi \times q} \mapsto \bR^{n_\Psi \times n_\Psi \times q}$. The activation of $\Psi$ is conducted element-wise as:
\begin{equation}
\begin{aligned}
	\mathcal{A}_{\rb{i}} \bb{x_1, x_2} & = \alpha\left(\Psi_{(i)}[x_1,x_2]\right),
\end{aligned}
\end{equation}
where $\alpha: \bR \mapsto \bR$ is an activation function (a scalar function). Typical activation functions include the sigmoid, hyperbolic tangent (tanh), and Rectified Linear Unit (ReLU):
\begin{equation}
\begin{aligned}
	\alpha_{\text{sig}}\rb{z}  & =  \frac{1}{1 + e^{-z}} \\
	\alpha_{\text{tanh}}\rb{z} & =  \tanh\rb{z}          \\
	\alpha_{\text{ReLU}}\rb{z} & = \max\rb{0, z},
\end{aligned}
\end{equation}

\begin{figure}[!htb]
	\begin{center}
		\includegraphics[width=1\linewidth]{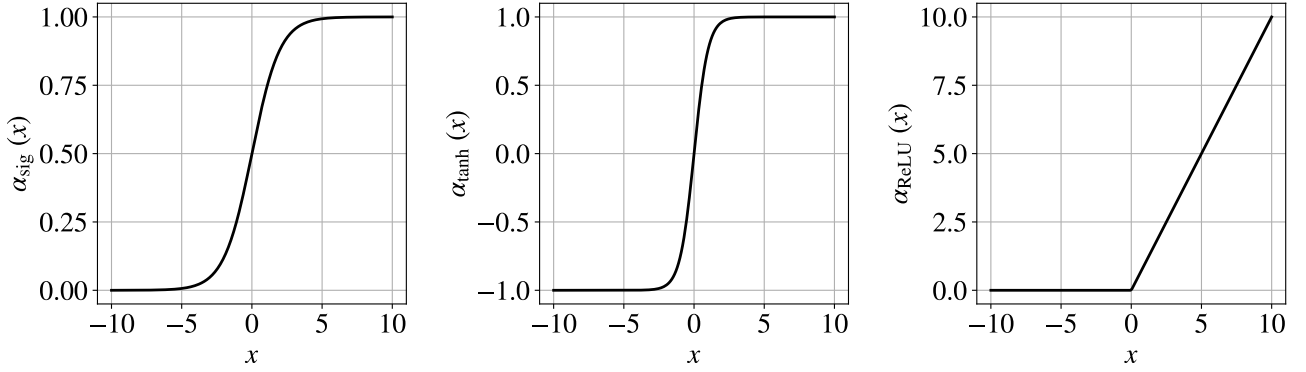}
		\caption{Output of the sigmoid, tanh and ReLU functions on a fixed domain.}
		\label{fig:activation}
	\end{center}
\end{figure}

\begin{figure}[!htb]
	\begin{center}
		\includegraphics[width=0.7\linewidth]{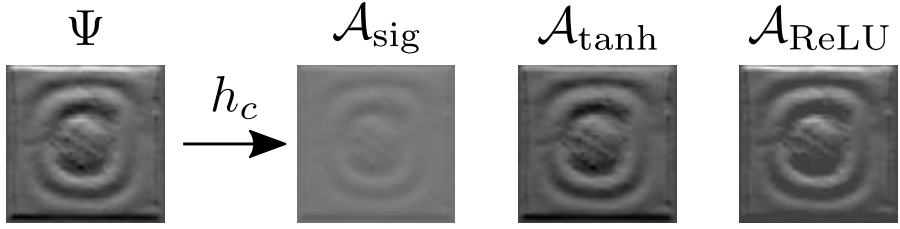}
		\caption{Element-wise activation of matrix $\Psi$ (a convolved micrograph) in the convolutional block.}
		\label{fig:activation2}
	\end{center}
\end{figure}

Figures~\ref{fig:activation} and \ref{fig:activation2} illustrate the transformation induced by the activation functions. These functions act as basis functions that, when combined in the CNN, enable capturing nonlinear behavior.  A common problem with sigmoid and tanh functions is that they exhibit the so-called {\it  vanishing gradient} effect  \cite{goodfellow2016deep}. Specifically, when the input values are large or small in magnitude, the gradient of the sigmoid and tanh functions is small (flat at both ends) and this makes the activation output insensitive to changes in the input. Furthermore, both the sigmoid and tanh functions are  sensitive to the change in input when the output is close to 1/2 and zero, respectively. The ReLU function is commonly used to avoid vanishing gradient effects and to  increase sensitivity \cite{nair2010rectified}. This activation function outputs a value of zero when the input is less than or equal to zero, but outputs the input value itself when the input is greater than zero. The function is linear when the input is greater than zero, which makes the CNN easier to optimize with gradient-based methods \cite{goodfellow2016deep}. However, ReLU is not continuously differentiable when the input is zero; in practice, CNN implementations assume that the gradient is zero when the input is zero.
\\

\subsection{Pooling Block}
\begin{figure}[htb!]
	\begin{center}
	\includegraphics[width=0.3\linewidth]{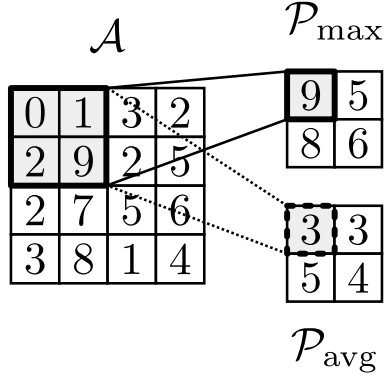}
	\caption{Max-pooling and average-pooling operations with a 2$\times$2 pooling window size. Each entry in the output feature map $\mathcal{P} \in \bR^{2 \times 2}$ is the maximum (or average) value of the corresponding pooling window in the input $\mathcal{A} \in \bR^{2 \times 2}$.}
	\label{fig:pool_layer}
	\end{center}
\end{figure}

The pooling block is a transformation that reduces the size of the output obtained by convolution and subsequent activation. This block also seeks to make the representation approximately invariant to small translation of the inputs \cite{goodfellow2016deep}. Max-pooling and average-pooling are the most common pooling operations (here we focus on max-pooling). The pooling operation $\mathcal{P} = f_p \rb{\mathcal{A}}$ can be expressed as a mapping $f_{p}: \bR^{n_\Psi \times n_\Psi \times q} \mapsto \bR^{\floor*{n_\Psi/n_p} \times \floor*{n_\Psi/n_p} \times q}$ and delivers a tensor $\mathcal{P}\in \bR^{\floor*{n_\Psi/n_p} \times \floor*{n_\Psi/n_p} \times q}$. To simplify the discussion, we denote $n_\mathcal{P} = \floor*{n_\Psi/n_p}$; the max-pooling operation with $i\in\{1, q\}$ is defined as:
\begin{equation}
\begin{aligned}
	\mathcal{P}_{\rb{i}}\bb{x_1, x_2} & = \text{max}\{\mathcal{A}_{\rb{i}}\bb{\rb{x_1-1}n_\mathcal{U}+n_\mathcal{U}', \rb{x_2-1}n_\mathcal{U}+n_\mathcal{U}'}, \; n_\mathcal{U}' \in \{1, n_\mathcal{U}\},\; n_\mathcal{U}' \in \{1, n_\mathcal{U}\}\},
\end{aligned}
\end{equation}
where $x_1\in\left\{1, n_\mathcal{P}\right\}$, $x_2\in\left\{1, n_\mathcal{P}\right\}$. In Figure~\ref{fig:pool_layer}, we illustrate the max-pooling and averaging pooling operations on a 1-channel input.
\\

Operations of convolution, activation, and pooling constitute a {\it  convolution unit} and deliver an output object $\mathcal{P}=f_p(h_c(f_c(\mathcal{V}))$. This object can be fed into another convolution unit to obtain a new output object $\mathcal{P}=f_p(h_c(f_c(\mathcal{P}))$ and this recursion can be performed over multiple units. The recursion has the effect of highlighting different features of the input data object $\mathcal{V}$ (e.g., that capture local and global patterns). In our discussion we consider a single convolution unit.

\subsection{Flattening Block}
The convolution, activation, and pooling blocks deliver a tensor $\mathcal{P} \in \bR^{n_\mathcal{P} \times n_\mathcal{P} \times p}$ that is flattened to a vector $v \in \bR^{n_\mathcal{P} \cdot n_\mathcal{P} \cdot p}$ {\color{black}(with some abuse of notation where $v$ was originally defined for input signal in 1D)}; this vector is fed into a fully-connected (dense) block that performs the final prediction. The vector $v$ is typically known as the {\it  feature vector}. The flattening block is represented as the mapping $f_f: \bR^{n_\mathcal{P} \times n_\mathcal{P} \times p} \mapsto \bR^{n_v}$ with $n_v = n_\mathcal{P} \cdot n_\mathcal{P} \cdot p$ that outputs $v = f_f\rb{\mathcal{P}}$. Note that this block is simply a vectorization of a tensor.

\begin{figure}[htb!]
	\begin{center}
		\includegraphics[width=0.5\linewidth]{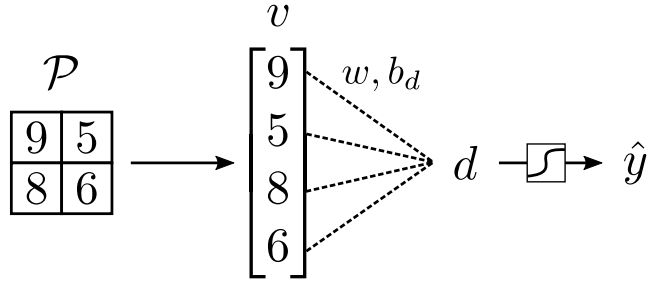}
		\caption{After flattening $\mathcal{P}$ to a vector $v$, the flatting block maps $v$ to $d$. The lines between $v$ and $d$ represent the dense connectivity induced by the weight vector $w$. A bias  $b_d$ is added after weighting. The prediction $\hat{y}$ is obtained by activating $d$.}
		\label{fig:dense_layer}
	\end{center}
\end{figure}

\subsection{Prediction (Dense) Block}
The feature vector $v \in \bR^{n_v}$ is input into a prediction block that delivers the final prediction $\hat{y}\in \mathbb{R}$. This block is typically a fully-connected (dense) neural net with multiple weighting and activation units (here we only consider a single unit). The prediction block first mixes the elements of the feature vector as:
\begin{equation}
\begin{aligned}
	d & = w^T v + b_d,
\end{aligned}
\end{equation}
where $w \in \bR^{ n_v}$ is the weighting (parameter) vector and $b_d \in \bR$ is the bias parameter. The {\color{black} scalar} $d\in \mathbb{R}$ is normally known as the {\it  evidence}. We denote the weighting operation using the mapping $f_d: \bR^{n_v} \mapsto \bR$ and thus $d=f_d(v; w,b_d)$. In Figure~\ref{fig:dense_layer}, we illustrate this weighting operation. An activation function is applied to the evidence $d$ to obtain the final prediction $\hat{y} = h_d (d)$ with $h_d: \bR \mapsto \bR$. This involves an activation (e.g., with a sigmoidal or ReLU function). We thus have that the prediction block has the form $\hat{y}=h_d(f_d(v))$. One can build a multi-layer fully-connected network by using a recursion of weighting and activation steps (here we consider one layer to simplify the exposition).  Here, we assume that the CNN delivers a single (scalar) output $\hat{y}$. In practice, however, a CNN can also predict multiple outputs; in such a case, the weight parameter becomes a matrix and the bias is a vector.  We also highlight that the predicted output can be an integer value (to perform classification tasks) or a continuous value (to perform regression tasks). 

\section{CNN Training}\label{sec:cnn_training}
CNN training aims to determine the parameters (operators and biases) that best map the input data to the output data. The fundamental operations in CNN training are forward and backward propagation, which are used by an optimization algorithm to improve parameters. In forward propagation, the input data is propagated forward through the CNN (block by block and layer by layer) to make a prediction and evaluate a loss function that captures the mismatch between the predicted and true output. In backward propagation, one computes the gradient of the loss function with respect to each parameter via a recursive application of the chain rule of differentiation (block by block layer by layer). This recursion starts in the prediction block and proceeds backwards to the convolution block. Notably, we will see that the derivatives of the transformation blocks have explicit (analytical) representations.  

\subsection{Forward Propagation}
To explain the process of forward propagation, we consider the CNN architecture shown in Figure~\ref{fig:architecture} with an input $\bV \in \bR^{n_\mathcal{V} \times n_\mathcal{V} \times p}$ and an output (prediction) $\hat{y} \in \bR$. All parameters $\bU \in \bR^{n_\mathcal{U} \times n_\mathcal{U} \times p \times q}$, $b_c \in \bR^{q}$, $w \in \bR^{n_v}$ and $b_d \in \bR$ are flattened and concatenated to form a parameter vector $\theta \in \bR^{n_{\theta}}$, where $n_{\theta} =n_\mathcal{U} \cdot n_\mathcal{U} \cdot p\cdot q + q + n_{\mathcal{V}} + 1$. We can express the entire sequence of operations carried out in the CNN as a {\it  composite mapping} $F: \bR^{n_\mathcal{V} \times n_\mathcal{V} \times p} \mapsto \bR$ and thus $\hat{y} = F\rb{\bV; \theta}$.  We call this   mapping the {\it  forward propagation} mapping (shown in Figure~\ref{fig:architecture}) and note that this can be decomposed sequentially (via lifting) as:
\begin{equation}
\begin{aligned}
	\hat{y} & = F\rb{\bV; \theta}                                                            \\
	        & = F\rb{\bV; \bU, b_c, w, b_d}                                                  \\
	        & =h_d \rb{f_d \rb{f_f \rb{f_{p} \rb{h_c \rb{f_c\rb{\bV; \bU, b_c}}}}; w, b_d}}.
\end{aligned}
\end{equation}
This reveals the nested nature and the order of the CNN transformations on the input data.  Specifically, we see that one conducts convolution, activation, pooling, flattening, and prediction.
\\

To perform training,  we collect a set of input samples $\bV^{(i)}\in \bR^{n_\mathcal{V} \times n_\mathcal{V} \times p}$ and corresponding output labels $y[i]\in \mathbb{R}$ with sample index $i\in\{1,n\}$.  Our aim is to compare the true output labels $y[i]$ to the CNN predictions $\hat{y}[i]= F\rb{\bV^{(i)}; \theta}$. To do so, we define the loss function $\mathcal{L}:\mathbb{R}^n\to \mathbb{R}$ that we aim to minimize. For classification tasks, the output $\hat{y}[i]$ is binary and a common loss function is the cross-entropy \cite{janocha2017loss}: 
\begin{equation}
\begin{aligned}
	\bL \rb{\hat{y}} & = -\frac{1}{n}\sum_{i=1}^{n}\bb{ y[i]\log{\rb{\hat{y}[i]}} + \rb{1-y[i]}\log\rb{1-\hat{y}[i]}},
\end{aligned}
\end{equation}
 For regression tasks, we have that the output $\hat{y}^{(i)}$ is continuous and a common loss function is the mean squared error: 
\begin{equation} 
\begin{aligned}
	\bL \rb{\hat{y}} = \frac{1}{n}\sum_{i=1}^{n}\rb{y[i] - \hat{y}[i] }^{2}.
\end{aligned}
\end{equation}
In compact form, we can write the loss function as $\bL \rb{F\rb{\bV; \theta}}=\bL \rb{\theta}$.  Here, $\mathcal{V}$ contains all the input samples $\mathcal{V}^{(i)}$ for $i\in \{1,n\}$. The loss function can be decomposed into individual samples and thus we can express it as:
\begin{equation} 
\begin{aligned}
	\bL \rb{\hat{y}} = \frac{1}{n}\sum_{i=1}^{n}\mathcal{L}^{(i)}(\theta). 
\end{aligned}
\end{equation}
where $\mathcal{L}^{(i)}(\theta)$ is the contribution associated with the $i$-th data sample.

\subsection{Optimization Algorithms}
The parameters of the CNN are obtained by finding a solution to the optimization problem:
\begin{equation}
\begin{aligned}
\min_\theta\; \bL  (\theta).
\end{aligned}
\end{equation}
The loss function $\bL$ embeds the forward propagation mapping $F$ (which is a highly nonlinear  mapping); as such, the loss function might contain many local minima (and such minima tend to be flat {\color{black}\cite{haeffele2017global}}). We recall that appearance of flat minima is often the manifestation of model overparameterization; this is due to the fact that the parameter vector $\theta$ can contain hundreds of thousands to millions of values. As such, training of CNNs need to be equipped with appropriate validation procedures (to ensure that the model generalizes well).  Most optimization algorithms used for CNNs are gradient-based (first-order); this is because second-order optimization methods (e.g., Newton-based) involve calculating the Hessian of the loss function $\bL$ with respect to parameter vector $\theta$, which is computationally expensive (or intractable).  Quasi-Newton methods (e.g., limited-memory BFGS) are also often used to train CNNs. 
\\

Gradient descent (GD) is the most commonly used algorithm to train CNNs.  This algorithm updates the parameters as:
\begin{equation}
\theta \leftarrow \theta - \eta\cdot \Delta \theta
\end{equation}
where $\eta\in \mathbb{R}_+$ is the learning rate (step length) and $\Delta \theta\in \mathbb{R}^{n_\theta}$ is the step direction. In GD, the step direction is set to $\Delta \theta=\nabla_{\theta}\mathcal{L}(\theta)$ (the gradient of the loss). Since the loss function is a sum over samples, the gradient can be decomposed as:
\begin{align}
\nabla_{\theta} \bL\rb{\theta}=\frac{1}{n}\sum_{i=1}^n\nabla_{\theta}\mathcal{L}^{(i)}(\theta).
\end{align}
Although GD is easy to compute and implement, $\theta$ is not updated until the gradient of the entire dataset is calculated (which can contain millions of data samples). In other words, the parameters are not updated until all the gradient components $\nabla_{\theta}\mathcal{L}^{(i)}(\theta)$ are available. This leads to load-imbalancing issues and limits the parallel scalability of the algorithm. 
\\

Stochastic gradient descent (SGD) is a variant of GD that updates $\theta$ more frequently (with gradients from a partial number of samples). Specifically, $\theta$ changes after calculating the loss for each sample. SGD updates the parameters by using the step $\Delta \theta=\nabla_{\theta}\mathcal{L}^{(i)}(\theta)$, where sample $i$ is selected at random. Since $\theta$ is updated frequently, SGD requires less memory and has better parallel scalability \cite{keuper2015asynchronous}. Moreover, it has been shown empirically that this algorithm has the tendency to escape local minima (it explores the parameter space better). However, if the training sample has high variance, SGD will converge slowly. 
\\

Mini-batch GD is an enhancement of SGD; here, the entire dataset is divided into $b$ batches and $\theta$ is updated after calculating the gradient of each batch. This results in a step direction of the form:
\begin{equation}
\Delta \theta= \frac{1}{b} \sum_{j\in \mathcal{B}^{(i)}} \nabla_{\theta} \bL^{(j)}\rb{\theta}.
\end{equation}
where $\mathcal{B}^{(i)}$ is a set of sample corresponding to batch $i\in \{1,b\}$ and the entries of the batches are selected at random. Mini-batch GD updates the model parameters frequently and has faster convergence in the presence of high variance \cite{wang2013variance}. 

\subsection{Backward Propagation}
Backward propagation (backpropagation) seeks to compute elements of the gradient of the loss $\nabla_{\theta} \mathcal{L}^{(i)}(\theta)$ by recursive use of the chain rule. This approach exploits the nested nature of the forward propagation mapping:
\begin{align}
	\hat{y} & =h_d \rb{f_d \rb{f_f \rb{f_{p} \rb{h_c \rb{f_c\rb{\bV; \bU, b_c}}}}; w, b_d}}.
\end{align}
This mapping can be expressed in backward form as:
\begin{subequations}
\begin{align}
	\hat{y}     & =h_d(d)                           \\
	d           & =f_d(v;w,b_d)                     \\
	v           & =f_f(\mathcal{P})                 \\
	\mathcal{P} & =f_p(\mathcal{A})                 \\
	\mathcal{A} & =h_c(\Psi)                        \\
	\Psi        & =f_c(\mathcal{V};\mathcal{U},b_c)
\end{align}
\end{subequations}
An illustration of this backward sequence is presented in Figure~\ref{fig:architecture}. Exploiting this structure is essential to enable scalability (particularly for saving memory). To facilitate the explanation, we consider the squared error loss defined for a single sample (generalization to multiple samples is trivial due to the additive nature of the loss): 
\begin{equation}
\bL \rb{\theta} = \frac{1}{2}(\hat{y} -y)^{2}.
\end{equation}
Our goal is to compute the gradient (the parameter step direction) $\Delta \theta=\nabla\mathcal{L}(\theta)$; here, we have that parameter $\theta=(w,b_d,\mathcal{U},b_c)$. 

\paragraph{Prediction Block Update.}  The step direction for the parameters $w$ (the gradient) in the prediction block is defined as $\Delta w \in \bR^{n_v}$ and its entries are given by:
\begin{equation}
\begin{aligned}
	\Delta w\bb{i} & = \frac{\partial \bL}{\partial w\bb{i}} \\
	               & = \frac{\partial \bL}{\partial \hat{y}} \frac{\partial \hat{y}}{\partial w\bb{i}} \\
	               & = \rb{\hat{y}-y} \cdot \frac{\partial \alpha\rb{d}}{\partial d}\frac{\partial d}{\partial w\bb{i}} \\
	               & = \rb{\hat{y}-y} \cdot \hat{y} \cdot \rb{1-\hat{y}} \cdot \frac{\partial \left(\displaystyle \sum_{i=1}^{n_v}w\bb{i}\cdot v\bb{i} + b_d\right)}{\partial w\bb{i}} \\
	               & =  \rb{\hat{y}-y} \cdot \hat{y} \cdot \rb{1-\hat{y}} \cdot v\bb{i},
\end{aligned}
\end{equation}
with $i\in \{1,n_v\}$. If we define $\Delta \hat{y} = \rb{\hat{y}-y} \cdot \hat{y} \cdot \rb{1-\hat{y}}$ then we can establish that the update has the simple form $\Delta w\bb{i} = \Delta \hat{y}\cdot v\bb{i} $ and thus: 
\begin{align}
\Delta w = \Delta \hat{y} \cdot v.
\end{align}
The gradient for the bias $\Delta b_d\in \mathbb{R}$ is given by:
\begin{equation}
\begin{aligned}
	\Delta b_d & = \frac{\partial \bL}{\partial b_d}  \\
	           & = \frac{\partial \bL}{\partial \hat{y}} \frac{\partial \hat{y}}{\partial b_d}  \\
	           & = \rb{\hat{y}-y} \cdot \hat{y} \cdot \rb{1-\hat{y}} \cdot \frac{\partial \left(\displaystyle \sum_{i=1}^{n_v} w\bb{i}\cdot v\bb{i} + b_d\right)}{\partial b_d} \\
	           & = \rb{\hat{y}-y} \cdot \hat{y}\cdot \rb{1-\hat{y}},
\end{aligned}
\end{equation}
and thus:
\begin{equation}
\Delta b_d = \Delta \hat{y}.
\end{equation}

\paragraph{Convolutional Block.} An important feature of CNN is that  some of its parameters (the convolution operators) are tensors and we thus require to compute gradients with respect tensors. While this sounds complicated, the gradient of the convolution operator $\Delta \bU$ has a remarkably intuitive structure (provided by the grid nature of the data and the structure of the convolution operation). To see this, we begin the recursion at:
\begin{equation}
v = f_f\rb{f_{p}\rb{h_c\rb{f_c\rb{\bV; \bU, b_c}}}}.
\end{equation}
Before computing $\Delta \bU_{(i,j)}$, we first obtain the gradient for the feature vector $\Delta v$ as:
\begin{equation}
\begin{aligned}
	\Delta v\bb{i} & = \frac{\partial \bL}{\partial v\bb{i}} \\
	               & = \frac{\partial \bL}{\partial \hat{y}} \cdot \frac{\partial \hat{y}}{\partial v \bb{i}} \\
	               & =  \rb{\hat{y}-y} \cdot \hat{y} \cdot \rb{1-\hat{y}} \cdot \frac{\left(\displaystyle \partial \sum_{i=1}^{n_v}w\bb{i}\cdot v\bb{i} + b_d\right)}{\partial v\bb{i}} \\
	               & = \rb{\hat{y}-y} \cdot \hat{y} \cdot \rb{1-\hat{y}} \cdot w\bb{i} \\
	               & = \Delta \hat{y} \cdot w\bb{i},
\end{aligned}
\end{equation}
with $i\in \{1,n_v\}$ and thus:
\begin{equation}
\Delta v = \Delta \hat{y}\cdot w.
\end{equation}
If we define the inverse mapping of the flattening operation as $f_f^{-1}: \bR^{n_v} \mapsto \bR^{n_\mathcal{P} \times n_\mathcal{P} \times q}$, we can express the update of the pooling block as:
\begin{equation}
\Delta \mathcal{P} = f_f^{-1}\rb{\Delta v}.
\end{equation}
The gradient from the pooling block $\Delta \mathcal{P}$ is passed to the activation block via up-sampling. If we have applied max-pooling with a $n_p \times n_p$ pooling operator, then:
\begin{equation}
\begin{aligned}
\Delta \mathcal{A}_{\rb{j}}\bb{x_1,x_2} =
	\begin{cases}
		\Delta \mathcal{P}_{\rb{j}}\bb{x_1/n_p, x_2/n_p}, & \text{if } x_1 \bmod n_p = 0 \text{ and } x_2 \bmod n_p = 0 \\
		0,                                                & \text{otherwise},
	\end{cases}
\end{aligned}
\end{equation}
where $a \mathrm{~mod~} b$ calculates the remainder obtained after dividing $a$ by $b$.
\\

The gradient with respect to the convolution operator has entries of the form:
\begin{equation}
\begin{aligned}
	\Delta \bU_{(i,j)}\bb{x_1,x_2} & = \frac{\partial \bL}{\partial \bU_{(i,j)}\bb{x_1,x_2}} \\
	                               & = \sum_{x_1'=1}^{n_\Psi}\sum_{x_2'=1}^{n_\Psi}\frac{\partial \bL}{\partial \mathcal{A}_{\rb{j}}\bb{x_1', x_2'}} \cdot \frac{\partial \mathcal{A}_{\rb{j}}\bb{x_1', x_2'}}{\partial \Psi_{\rb{j}}\bb{x_1',x_2'}} \cdot \frac{\partial \Psi_{\rb{j}}\bb{x_1',x_2'}}{\partial \bU_{(i,j)}\bb{x_1,x_2}} \\
	                               & =  \sum_{x_1'=1}^{n_\Psi}\sum_{x_2'=1}^{n_\Psi} \Delta \mathcal{A}_{\rb{j}}\bb{x_1', x_2'} \cdot \mathcal{A}_{\rb{j}}\bb{x_1', x_2'} \cdot \rb{1-\mathcal{A}_{\rb{j}}\bb{x_1', x_2'}} \cdot \frac{\partial \Psi_{\rb{j}}\bb{x_1', x_2'}}{\partial \bU_{(i,j)}\bb{x_1,x_2}}
\end{aligned}
\end{equation}
for $i\in\{1,p\}$ and $j\in \{1,q\}$, and with:
\begin{equation}
\begin{aligned}
	\frac{\partial \Psi_{\rb{j}}\bb{x_1',x_2'}}{\partial \bU_{(i,j)}\bb{x_1,x_2}} & =  \frac{\partial \rb{b_{c}[j]+\displaystyle\sum_{i=1}^{p}  \bU_{(i,j)}\bb{x_1,x_2} \bV_{\rb{i}}\bb{x_1+x_1'-1,x_2+x_2'-1} }}{\partial \bU_{(i,j)}\bb{x_1,x_2}} \\
	                                                                              & =  \bV_{\rb{i}}\bb{x_1+x_1'-1,x_2+x_2'-1}.
\end{aligned}
\end{equation}
If we define $\Delta \Psi_{\rb{j}}\bb{x_1', x_2'} = \Delta \mathcal{A}_{\rb{j}}\bb{x_1', x_2'} \cdot \mathcal{A}_{\rb{j}}\bb{x_1', x_2'} \cdot \rb{1-\mathcal{A}_{\rb{j}}\bb{x_1', x_2'}}$, then:
\begin{equation}
\begin{aligned}
	\Delta \bU_{\rb{i,j}}\bb{x_1,x_2} & =\sum_{x_1'=1}^{n_\Psi}\sum_{x_2'=1}^{n_\Psi} \Delta \Psi_{\rb{j}}\bb{x_1', x_2'} \cdot \bV_{\rb{i}}\bb{x_1+x_1'-1,x_2+x_2'-1}.
\end{aligned}
\end{equation}
\\

The gradient with respect to the bias parameters in the convolution block has entries:
\begin{equation}
\begin{aligned}
	\Delta b_{c}[j] & = \frac{\partial \bL}{\partial b_{c}[j]} \\
	                & = \sum_{x_1'=1}^{n_\Psi} \sum_{x_2'=1}^{n_\Psi} \frac{\partial \bL}{\partial \mathcal{A}_{\rb{j}}\bb{x_1', x_2'}}\cdot \frac{\partial \mathcal{A}_{\rb{j}}\bb{x_1', x_2'}}{\partial \Psi_{\rb{j}}\bb{x_1', x_2'}}\cdot \frac{\partial \Psi_{\rb{j}}\bb{x_1', x_2'}}{\partial b_{c}[j]} \\
	                & = \sum_{x_1'=1}^{n_\Psi} \sum_{x_2'=1}^{n_\Psi} \Delta \mathcal{A}_{\rb{j}}\bb{x_1', x_2'} \cdot \mathcal{A}_{\rb{j}}\bb{x_1', x_2'} \cdot \rb{1-\mathcal{A}_{\rb{j}}\bb{x_1', x_2'}} \cdot \frac{\partial \Psi_{\rb{j}}\bb{x_1',x_2'}}{\partial b_{c}[j]}
\end{aligned}
\end{equation}
with $j\in \{1,q\}$. Because $\Delta \Psi_{\rb{j}}\bb{x_1', x_2'} = \Delta \mathcal{A}_{\rb{j}}\bb{x_1', x_2'} \cdot \mathcal{A}_{\rb{j}}\bb{x_1', x_2'} \cdot \rb{1-\mathcal{A}_{\rb{j}}\bb{x_1', x_2'}}$, and
\begin{equation}
\begin{aligned}
	\frac{\partial \Psi_{\rb{j}}\bb{x_1',x_2'}}{\partial b_{c}[j]} & =  \frac{\partial \rb{b_{c}[j]+\sum_{i=1}^{p}  \bU_{(i,j)}\bb{x_1,x_2} \bV_{\rb{i}}\bb{x_1+x_1'-1,x_2+x_2'-1} }}{\partial b_{c}[j]} \\
	                                                               & =  1,
\end{aligned}
\end{equation}
we have that:
\begin{equation}
	\Delta b_{c}[j] = \sum_{x_1'=1}^{n_\Psi} \sum_{x_2'=1}^{xx} \Delta \Psi_{\rb{j}}\bb{x_1', x_2'}.
\end{equation}

\subsection{Saliency Maps}

Saliency maps is a useful technique used to determine features that best explain a prediction; in other words, these techniques seek to highlight what the CNN searches for in the data. The most basic version of the saliency map is based on gradient information for the loss (with respect to the input data) \cite{simonyan2013deep}. Recall that the forward propagation function is given by $F\rb{\bV; \theta}$, where the input is $\bV \in \bR^{n_\mathcal{V} \times n_\mathcal{V} \times p}$, and the parameter vector is $\theta$. We express the loss function as $\bL \rb{F\rb{\bV; \theta}}$ (for input sample $\bV$); the {\it  saliency map} is a tensor $\mathcal{S} \in \bR^{n_\mathcal{V} \times n_\mathcal{V} \times p}$ of the same dimension as the input $\bV$ and is given by: 
\begin{equation}
\mathcal{S} = \text{abs} \rb{\frac{\partial \bL \rb{F\rb{\bV; \theta}}}{\partial \bV}}.
\end{equation}
If we are only interested in the location of the most important regions, we can take the maximum value of $\mathcal{S}$ over all channels $p$ and generate the saliency {\it  mask} $S \in \bR^{n_\mathcal{V} \times n_\mathcal{V}}$ (a matrix). In some cases, we can also study the saliency map for each channel, as this may have physical significance. Saliency maps are easy to implement (can be obtained via back-propagation) but suffer from vanishing gradients effects. The so-called {\it  integrated gradient} (IG)  seeks to overcome vanishing gradient effects \cite{sundararajan2017axiomatic}; the IG is defined as:
\begin{equation}
\mathcal{S}_{IG} = \text{abs}\rb{\rb{\bV-\bar{\bV}} \cdot \int_{0}^{1} \frac{\partial \bL\rb{F\rb{\bar{\bV}+\beta \rb{\bV-\bar{\bV}}; \theta}}}{\partial \bV} d\beta},
\end{equation}
where $\bar{\bV}$ is a {\it  baseline input} that represents the absence of a feature in the input $\bV$. Typically, $\bar{\bV}$ is a sparse object (few non-zero entries). 

\section{Applications}\label{sec:cases}
CNNs have been applied to different problems arising in chemical engineering (and related fields) but have been applied mostly to image data. In this section, we present different case studies to highlight the broad versatility of CNNs; specifically, we show how to use CNNs to make predictions from grid data objects that are general vectors (in 1D), matrices (in 2D), and tensors (in 3D).  Our discussion also seeks to highlight the different concepts discussed. The scripts to reproduce the results for the optimal control (1D), the multivariate process monitoring (2D) and the molecular simulations (3D) case studies can be found here \url{https://github.com/zavalab/ML/tree/master/ConvNet}. 

\subsection{Optimal Control (1D)}
\begin{figure}[htb!]
	\begin{center}
		\includegraphics[width=1\linewidth]{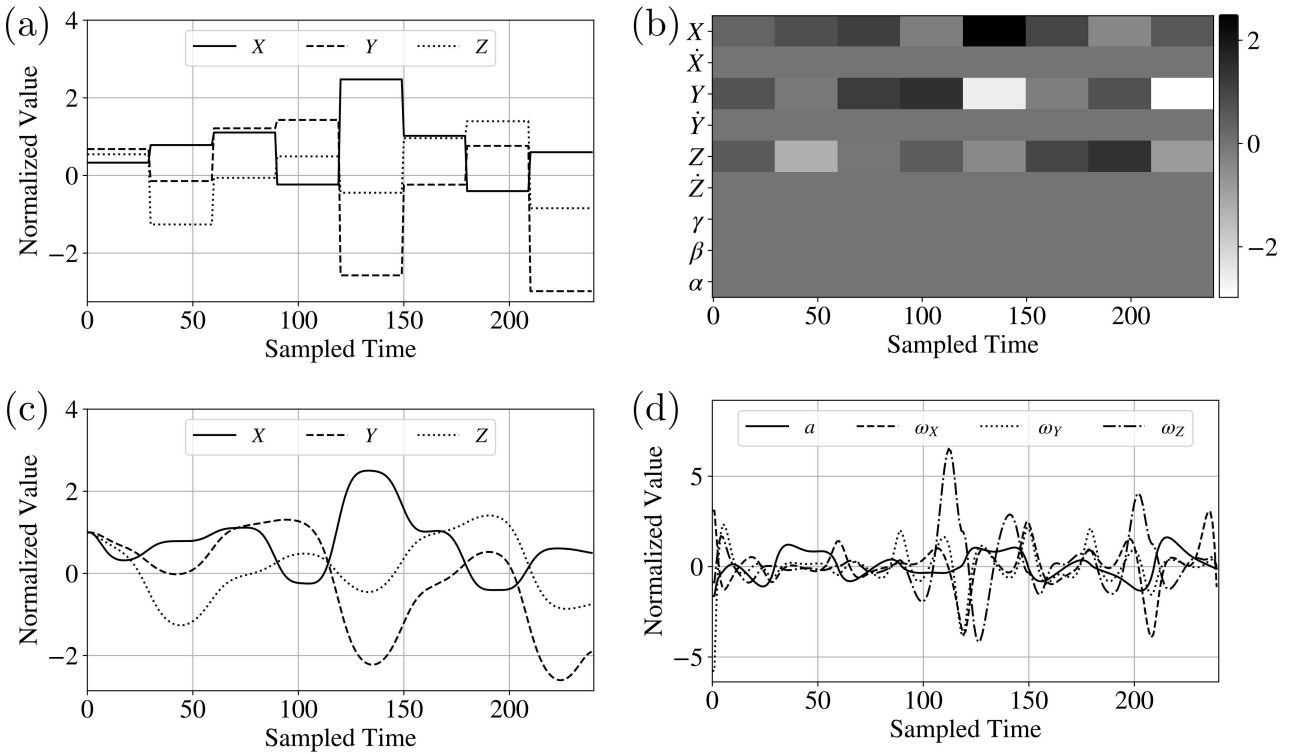}
		\caption{Grid representation of time-series data. (a) Reference time-series data of state variable $X, Y$ and $Z$. The variables are normalized with a zero mean and a unit variance. Other reference state variables $\dot{X}, \dot{Y}, \dot{Z}, \gamma, \beta$ and $\alpha$ are all zero. (b) A ${240 \times 9}$ image, where the row dimension is the number of state variables and the column dimension is the number of time points. Each row of the image represents one of the time series in (a). (c) Optimal solution for state variables. (d) Optimal solutions for control variables.}
		\label{fig:rotor1}
	\end{center}
\end{figure}

Optimal control problems (OCP) are optimization problems that are commonly encountered in engineering. The basic idea is to find a control trajectory that drives a dynamical system in some optimal sense. A common goal, for instance, is to find a control trajectory that minimizes the distance of the system state to a given reference trajectory. Here, we study nonlinear OCPs for a quadrotor system  \cite{hehn2011flying}.  In this problem, we have an inverted pendulum that is placed on a quadrotor, which is required to fly along a reference trajectory. The governing equations of the quadrotor-pendulum system are:
\begin{equation}
\begin{aligned}
	\frac{d^2 X}{d t^2} &= a\rb{\cos \gamma \sin \beta \cos \alpha + \sin \gamma \sin \alpha} \\
	\frac{d^2 Y}{d t^2} &= a\rb{\cos \gamma \sin \beta \sin \alpha - \sin \gamma \cos \alpha} \\
	\frac{d^2 Z}{d t^2} &= a\cos \gamma \cos \beta - g \\
	\frac{d \gamma}{d t} &= \rb{\omega_X \cos \gamma + \omega_Y \sin \gamma} / \cos \beta \\
	\frac{d \beta}{d t} &= -\omega_X \sin \gamma + \omega_Y \cos \gamma \\
	\frac{d \alpha}{d t} &= \omega_X \cos \gamma \tan \beta + \omega_Y \sin \gamma \tan \beta + \omega_Z. 
\end{aligned}
\end{equation}
where $X, Y, Z$ are the positions and $\gamma, \beta, \alpha$ are angles. We define $v = (X, \dot{X}, Y, \dot{Y}, Z, \dot{Z}, \gamma, \beta, \alpha)$ as state variables, where $\dot{X}, \dot{Y}, \dot{Z}$ are velocities. The control variables are $u = (\alpha, \omega_X, \omega_Y, \omega_Z)$, where $\alpha$ is the vertical acceleration and $\omega_X, \omega_Y, \omega_Z$ are the angular velocities of the quadrotor fans. Our goal is to find a control trajectory that minimizes the cost:
\begin{equation}\label{eqn:rotor}
\begin{aligned}
	y = \min_{u,v} \; \int_{0}^T\frac{1}{2} \norm{v(t) - v^{\mathrm{ref}}(t)}^2 + \frac{1}{2} \norm{u(t)}^2dt,
\end{aligned}
\end{equation}
where $v^{\mathrm{ref}}$ is a time-varying reference trajectory (changes every 30 seconds) and $u^{\mathrm{ref}}$ is assumed to be zero. Minimization of this cost seeks to keep the quadrotor trajectory close to the reference trajectory with as little control action (energy) as possible. An example reference trajectory $v^{\mathrm{ref}}$ is shown in Figure~\ref{fig:rotor1}; here, we show the reference trajectories for $X, Y$ and $Z$ and the trajectories for the rest of the states are assumed to be zero. 
\\

In this case study, we want to train a 1D CNN to predict optimal cost values $y$ based on a given reference trajectory $v^{\mathrm{ref}}$ for the states (without the need of solving the OCP). This type of capability can be used to determine reference trajectories that are least/most impactful on performance (i.e., trajectories that are difficult or easy to track). {\color{black} CNNs can also be used to approximate optimal control laws, which can significantly reduce online computational time and enable fast model predictive control implementations.} To train the CNN, we generated 200 random trajectories $v^{\mathrm{ref}}$ and solved the corresponding OCPs. Each OCP was discretized, modeled with JuMP \cite{dunning2017jump}, and solved with IPOPT \cite{wachter2006implementation} \cite{bezanson2017julia}. After solving each OCP, we obtained the optimal states and controls $v$ and $u$ (shown in Figure~\ref{fig:rotor1}) and the optimal cost $y$. 

The cost is the output of the 1D CNN; to obtain the input, we need to represent the reference trajectory $v^{\mathrm{ref}}$ (which includes multiple state variables) as a 1D grid object that the CNN can analyze. Here, we represent $v^{\mathrm{ref}}$ as a multi-channel 1D grid object.  The data object is  denoted as $\mathcal{V}$; this object has $p=9$ channels and each channel $\mathcal{V}_{(i)}$ is a vector of dimension $n_\mathcal{V}=240$ (each channel contains the trajectory of one of the state variables).We visualize this multi-channel object as a grayscale image (each row is channel), see Figure~\ref{fig:rotor1}. We can see that this representation provides a different perspective on the nature of the reference trajectories (in the form of patterns). The goal is to train the 1D CNN to recognize how features of such patterns impact cost (e.g., abrupt vs. mild changes in the reference trajectory). We note that the data object $\mathcal{V}$ could, in principle, be represented as a single-channel matrix that could be fed to a 2D CNN. However, this approach would require 2D convolution operations (as opposed to the cheaper 1D convolutions). The results presented here seek to highlight that a 1D CNN suffices and gives accurate predictions. This also seeks to highlight the versatility that one has in representing data objects in different forms. 

\begin{figure}[!htb]
	\begin{center}
		\includegraphics[width=1\linewidth]{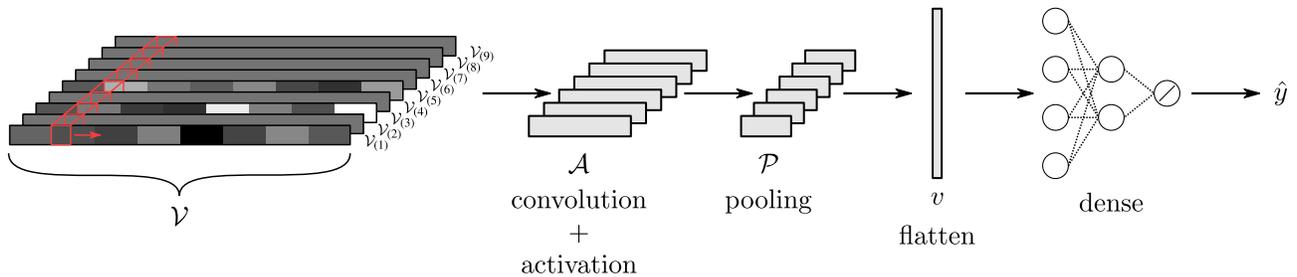}
		\caption{1D CNN architecture. The input to the 1D CNN is a 9-channel vector object $\mathcal{V}$, each channel $\mathcal{V}_{(i)}$ contains a vector of dimension 240 ($n_\mathcal{V}=240$ and $p=9$) The 1D CNN comprises a convolutional block that contains a collection of 64 operators $\mathcal{U}$; each operator is a vector of dimension 3  ($q=64$ and $n_\mathcal{U}=3$). The max-pooling block uses $n_p=2$. The red box illustrates the scanning of the 1D convolution operator (convolutions are applied to each of the 9 channels separately). The object $\mathcal{A}$ generated by the convolution and activation blocks contains 64 channels, each given by a vector of dimension 238  ($n_{\mathcal{A}}=238$ and $p=64$). The max-pooling block yields an object $\mathcal{P}$ with 64 channels, and each channel has a vector of dimension 119 ($n_{\mathcal{P}}=119$ and $p=64$). The pooling object is flattened into feature vector $v \in \bR^{7616}$; this vector is passed to the dense layers (each having 32 hidden units). The predicted cost $\hat{y} \in \bR$ is the output from the dense layer activated by a linear function.}
		\label{fig:rotor2}
	\end{center}
\end{figure}

\begin{figure}[!htb]
	\begin{center}
		\includegraphics[width=0.4\linewidth]{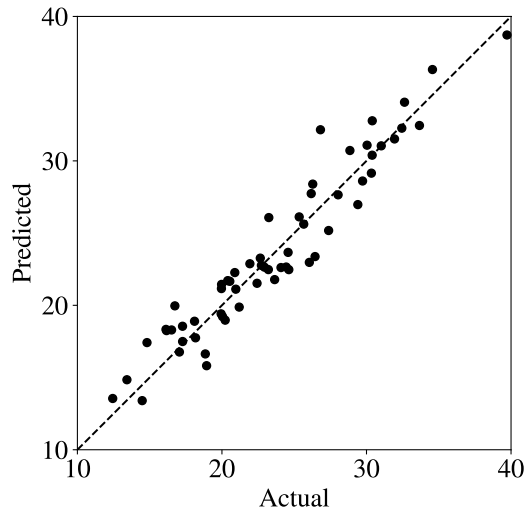}
		\caption{Predicted and actual cost from 1D CNN. The black dashed line represents a perfect correlation. The RMSE of cost prediction is 1.71.}
		\label{fig:rotor3}
	\end{center}
\end{figure}

The training:validation:testing data ratio used is 3:1:2. The 1D CNN architecture is of the form Conv(64)-MaxPool-Flatten-Dense(32)-Dense(32), where the number in the parentheses is either the number of convolution operators or dense layer units. The architecture is shown in Figure~\ref{fig:rotor2}; here, we can see that we have a single convolution unit but the prediction block has a couple of layers. The final prediction block generates a scalar prediction $\hat{y}$ (corresponding to the cost). Regression results for the 1D CNN are presented in Figure \ref{fig:rotor3}; we can see that accurate predictions can be obtained. {\color{black} Compared with a feed-forward neural net (RMSE=5.44) and a linear regression (RMSE=8.56), the 1D CNN has a more accurate prediction.} This performance is quite surprising given the complex nature of the reference trajectories and the nonlinearity of the system. Specifically, the 1D CNN has learned to recognize patterns that have the strongest effect on cost.

\subsection{Flow Cytometry (2D) }
This case study is based on the results presented in \cite{jiang2021endotoxin}; our goal here is not to revisit these results; instead, we seek to highlight the data representation aspects of the problem (how to start from raw data into an input representation that the CNN can process).  
\\

\begin{figure}[!htb]
	\begin{center}
		\includegraphics[width=1\linewidth]{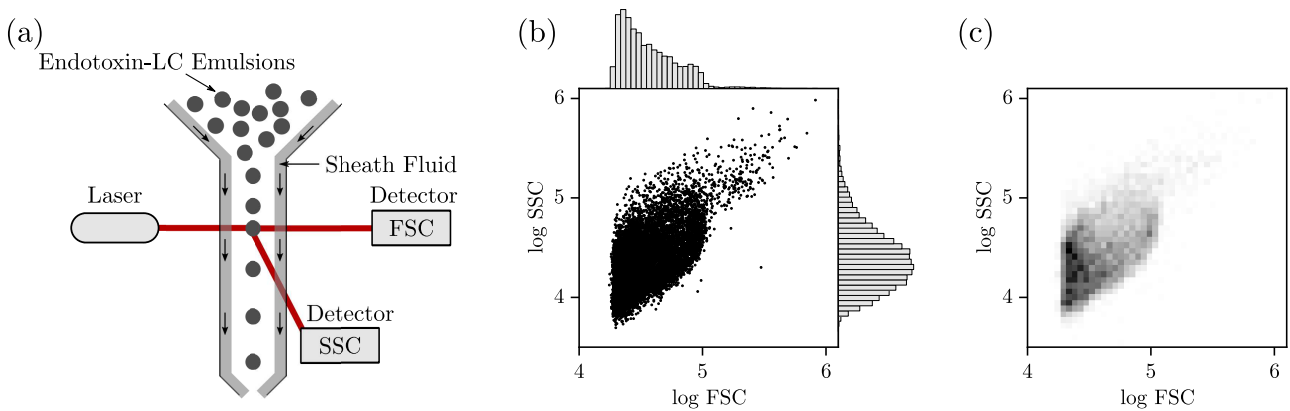}
		\caption{Overview of the interaction between endotoxin and LC emulsions. (a) Generation of FSC/SSC scatter fields. The endotoxin-LC emulsions are pumped into the flow cytometer in the direction of the sheath flow. Laser light is scattered from the LC droplets and collected at two angles (FSC and SSC). By combining the FSC and SSC data points for 10,000 LC droplets, we generate an FSC/SSC field. (b) Scatter field generated by LC droplets exposed to 100 pg/mL of endotoxin. Marginal probability densities of FSC (top) and SSC (right) light in log scale are generated with 50 bins. (c) 2D grid of the scatter fields by binning and counting the number of events in a bin. Reproduced from \cite{jiang2021endotoxin} with permission from the Royal Society of Chemistry.}
		\label{fig:endo1}
	\end{center}
\end{figure}

Endotoxins are lipopolysaccharides (LPS) present in the outer membrane of gram-negative bacteria; these agents are responsible for the pathophysiological phenomena associated with gram-negative infections, such as endotoxemia (septic shock caused by a severe immune response) \cite{makela1984handbook}. It has been recently discovered that micrometer-sized liquid crystal (LC) droplets dispersed in an aqueous solution can be used as a sensing method to detect and measure the concentration of endotoxins of different bacterial organisms. After exposure to endotoxins, LC droplets undergo ordering transitions that yield distinct optical properties and this change can be detected using flow cytometry (as shown in Figure~\ref{fig:endo1}). Flow cytometry produces scatter point clouds of forward and side scattering (FSC/SSC) as shown in Figure~\ref{fig:endo1} (each point represents a scattering event of a given droplet). A point could be converted into a 2D grid data object via binning; this is done by discretizing the FSC/SSC domain and by counting the number of points in a bin. The 2D grid object obtained is a matrix that we visualize as a grayscale image; here, each pixel is a bin and the intensity is the number of events in the bin (bins with more events appear darker). We note that this visualization is a 2D projection of a 3D histogram (the third dimension corresponds to the number of events, also known as the frequency). In other words, the 2D grid object captures the geometry (shape) of the joint probability density of the FSC/SSC. This shape has curvature and this can be characterized by gradients (derivatives). 
\\

\begin{figure}[!htb]
	\begin{center}
		\includegraphics[width=0.6\linewidth]{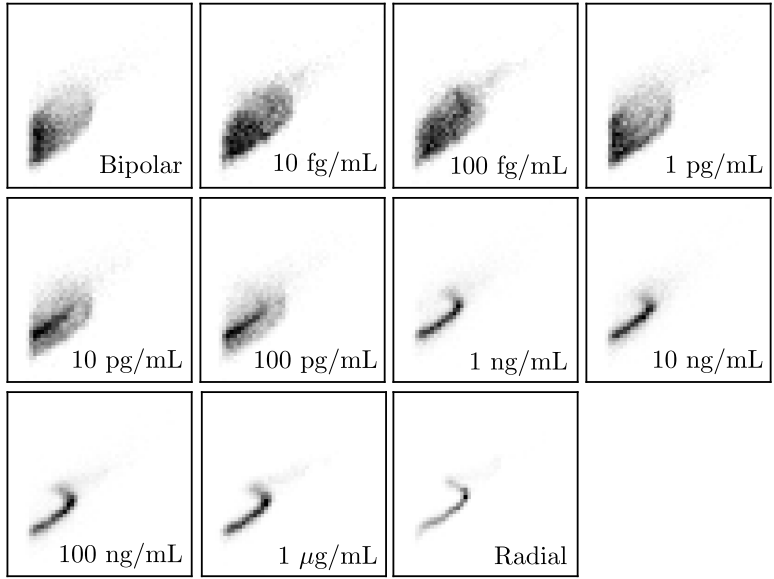}
		\caption{Effect of endotoxin concentration on the FSC/SSC scatter field (represented as 2D grid objects). Scatter fields obtained using LC droplets exposed to different concentrations of endotoxin. As the endotoxin concentration increases, the LC droplet population shifts from a bipolar to a radial configuration. Reproduced from \cite{jiang2021endotoxin} with permission from the Royal Society of Chemistry.}
		\label{fig:endo2}
	\end{center}
\end{figure}

The effect of endotoxin concentration on the FSC/SSC scatter fields (after binning) is shown in Figure~\ref{fig:endo2}; clear differences in the patterns can be observed at  concentrations that are far apart but difference are subtle for nearby concentrations.  Our goal is to train a 2D CNN that can predict concentrations from the scatter fields. 
\\ 

\begin{figure}[htb!]
	\begin{center}
		\includegraphics[width=1\linewidth]{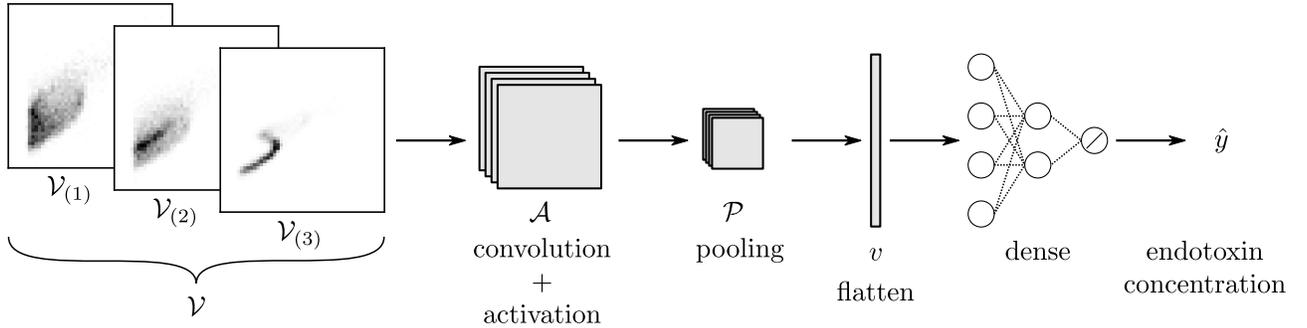}
		\caption{EndoNet architecture. The input to EndoNet is a 3-channel object $\mathcal{V} \in \bR^{50 \times 50 \times 3}$ ($n_\mathcal{V}=50$ and $p=3$); the channels correspond to the target, the negative reference, and the positive reference (each is a matrix of dimension $50 \times 50$). EndoNet includes a convolutional block with 64, $3 \times 3$ operators ($n_\mathcal{U}=3$ and $q=64$) and a max-pooling block with $2 \times 2$ operators ($n_p=2$). The feature map $\mathcal{A}$ generated by the convolution and activation blocks is a tensor $\bR^{48 \times 48 \times 64}$ ($n_\mathcal{A}=48$ and $p=64$). The max-pooling block generates a feature map $\mathcal{P} \in \bR^{24 \times 24 \times 64}$ ($n_\mathcal{P}=24$ and $p=64$) that is flattened into a long vector $v \in \bR^{36863}$. This vector is passed to two dense layers (each having 32 hidden units). The predicted endotoxin concentration $\hat{y} \in \bR$ is the output from the dense layer activated by a linear function.}
		\label{fig:endo3}
	\end{center}
\end{figure}

We used the following procedure to obtain the 2D grid data objects (samples) that were fed to the CNN. For each sample, we generated bins for a given scatter field by partitioning the FSC and SSC dimensions into 50 segments (the grid has $50 \times 50=2,500$ pixels). For each sample, we were also given reference scatter fields that represented limiting behavior: bipolar control (negative) and radial control (positive). Each sample is given by a 3-channel object $\bV$  where the first channel is the negative reference matrix $\bV_{\rb{1}}$, the second channel is the target matrix $\bV_{\rb{2}}$, and the third channel is a positive reference matrix $\bV_{\rb{3}}$ (each channel contains a $50\times 50$ matrix). This procedure is illustrated in Figure \ref{fig:endo3}. This multi-channel data representation approach has been found to magnify differences in the target matrix from the references (we will see that neglecting the negative/positive references does not give accurate predictions).  The data representation used highlights how multi-channel inputs can be used in creative ways to encode different properties of the data at hand.

\begin{figure}[!htb]
	\begin{center}
		\includegraphics[width=0.35\linewidth]{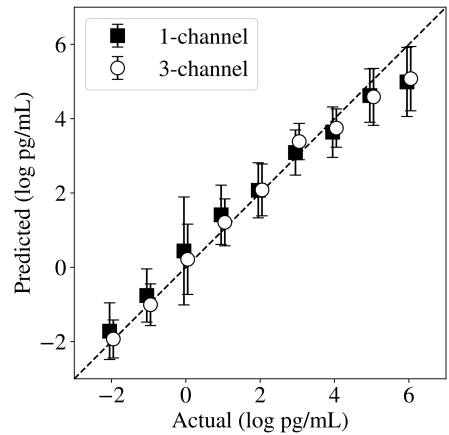}
		\caption{Predicted and actual concentrations at different concentrations using 1-channel and 3-channel data representation.}
		\label{fig:endo4}
	\end{center}
\end{figure}

The 3-channel data object was fed to a CNN, which we call {\it  EndoNet}. EndoNet has an architecture of Conv(64)-MaxPool-Flatten-Dense(32)-Dense(32)-Dense(1). The output block generates a scalar prediction $\hat{y}$ (corresponding to the endotoxin concentration). In other words, the CNN seeks to predict the endotoxin concentration from the input flow cytometry fields. The architecture of EndoNet is shown in Figure~\ref{fig:endo3}. The regression results for EndoNet are presented in Figure \ref{fig:endo4}. EndoNet can extract pattern information within and between each channel of the input image (which include negative and positive controls). Capturing differences between channels provides context for the CNN and has the effect of highlighting the domains in the scatter field that contain the most information. To validate this hypothesis, we conducted predictions for EndoNet using only the 1-channel representation as input (we ignored the positive and negative channels). For the 1-channel data representation we obtained an RMSE of 0.97; for the 3-channel representation we obtained and RMSE of 0.78. It is particularly remarkable that EndoNet can accurately predict concentrations that span eight orders of magnitude. 
\\

\begin{figure}[!htb]
	\begin{center}
		\includegraphics[width=0.7\linewidth]{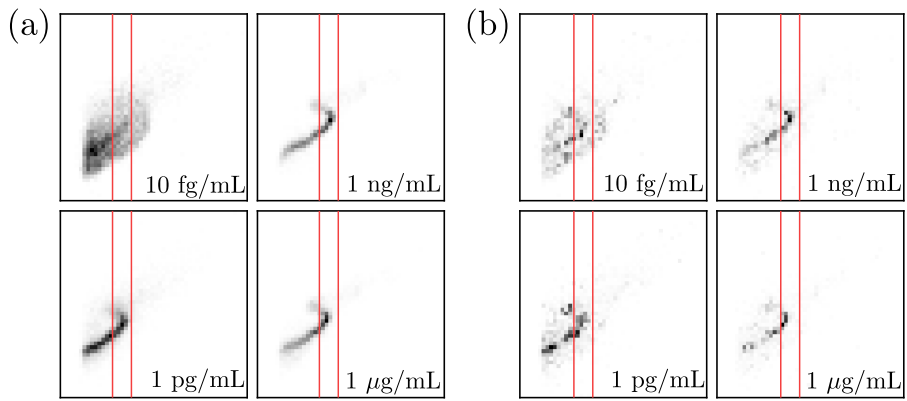}
		\caption{Saliency maps for analyzing important features. (a) The target channel matrix of the input $\bV$ to the EndoNet. (b) The corresponding saliency maps. The region between the two red lines is the characteristic S-region from the traditional counting method. The saliency maps highlight that the characteristic S-region contains significant information but that other regions are also relevant.}
		\label{fig:endo5}
	\end{center}
\end{figure}

We use saliency maps to highlight features that EndoNet is searching for to make predictions. The saliency map used is calculated by integrated gradients; here the map is a matrix $S \in \bR^{50 \times 50}$. Figure~\ref{fig:endo5} presents saliency maps obtained with EndoNet; here, we notice several important emerging patterns. First, the saliency maps clusters around a characteristic S-region used in traditional counting methods (the region between the red lines) \cite{miller2013analysis}. This indicates that EndoNet searches for information in the same region as this traditional method; however, the saliency maps also indicate that there is a clear diagonal pattern that must be considered. The saliency maps also reveal regions that do not provide information for classification and regression. Specifically, we see that regions of high FSC and SSC values do not provide significant information.

\subsection{Multivariate Process Monitoring (2D)}
Fault detection is a common task performed in process monitoring. Here, the idea is to collect multivariate time series (for different process variables) under different modes of operation (each mode is induced by a specific fault). The goal is to identify features (signatures) in the time series to determine if the process is a given mode. We focus on a case study for the Tennessee-Eastman (TE) process \cite{downs1993plant}. We expand the recent work presented in \cite{wu2018deep}, which shows that multivariate signals can be processed as matrices and used by CNNs to identify faults. This work is expanded by performing CNN architecture optimization and by performing saliency map analysis to identify critical variables that best explain faults.  
\\

The TE process units include a reactor, condenser, compressor, separator and stripper. The TE process produces two products (G and H) and a byproduct (F) from four reactants (A, C, D and E). Component B is an inert compound. In total, the TE process contains a total of 52 measured variables; 41 of them are process variables and 11 are manipulated variables. This process exhibits 20 different types of faults, as listed in Table~\ref{tab:case2fault}.
\\

\begin{table}[htb!]
	\centering
	\begin{threeparttable}[t]
		\caption{Types of Faults for TE Process \cite{downs1993plant, heo2018fault}.}
		\begin{tabular}{ccc}
			\toprule
			Fault ID   &                        Fault Name                        &       Type       \\ \midrule
			Fault 1   &    A/C feed ratio, B composition constant (stream 4)     &       Step       \\
			Fault 2   &       B composition, A/C ratio constant (stream 4)       &       Step       \\
			Fault 3   &              D feed temperature (stream 2)               &       Step       \\
			Fault 4   &         Reactor cooling water inlet temperature          &       Step       \\
			Fault 5   &        Condenser cooling water inlet temperature         &       Step       \\
			Fault 6   &                  A feed loss (stream 1)                  &       Step       \\
			Fault 7   & C header pressure loss - reduced availability (stream 4) &       Step       \\
			Fault 8   &           A, B, C feed composition (stream 4)            & Random variation \\
			Fault 9   &              D feed temperature (stream 2)               & Random variation \\
			Fault 10   &              C feed temperature (stream 4)               & Random variation \\
			Fault 11   &         Reactor cooling water inlet temperature          & Random variation \\
			Fault 12   &        Condenser cooling water inlet temperature         & Random variation \\
			Fault 13   &                    Reaction kinetics                     &    Slow drift    \\
			Fault 14   &               Reactor cooling water valve                &     Sticking     \\
			Fault 15   &              Condenser cooling water valve               &     Sticking     \\
			Fault 16-20 &                         Unknown                          &     Unknown      \\ \bottomrule
		\end{tabular}
		\label{tab:case2fault}
	\end{threeparttable}
\end{table}

\begin{figure}[htb!]
	\begin{center}
		\includegraphics[width=1\linewidth]{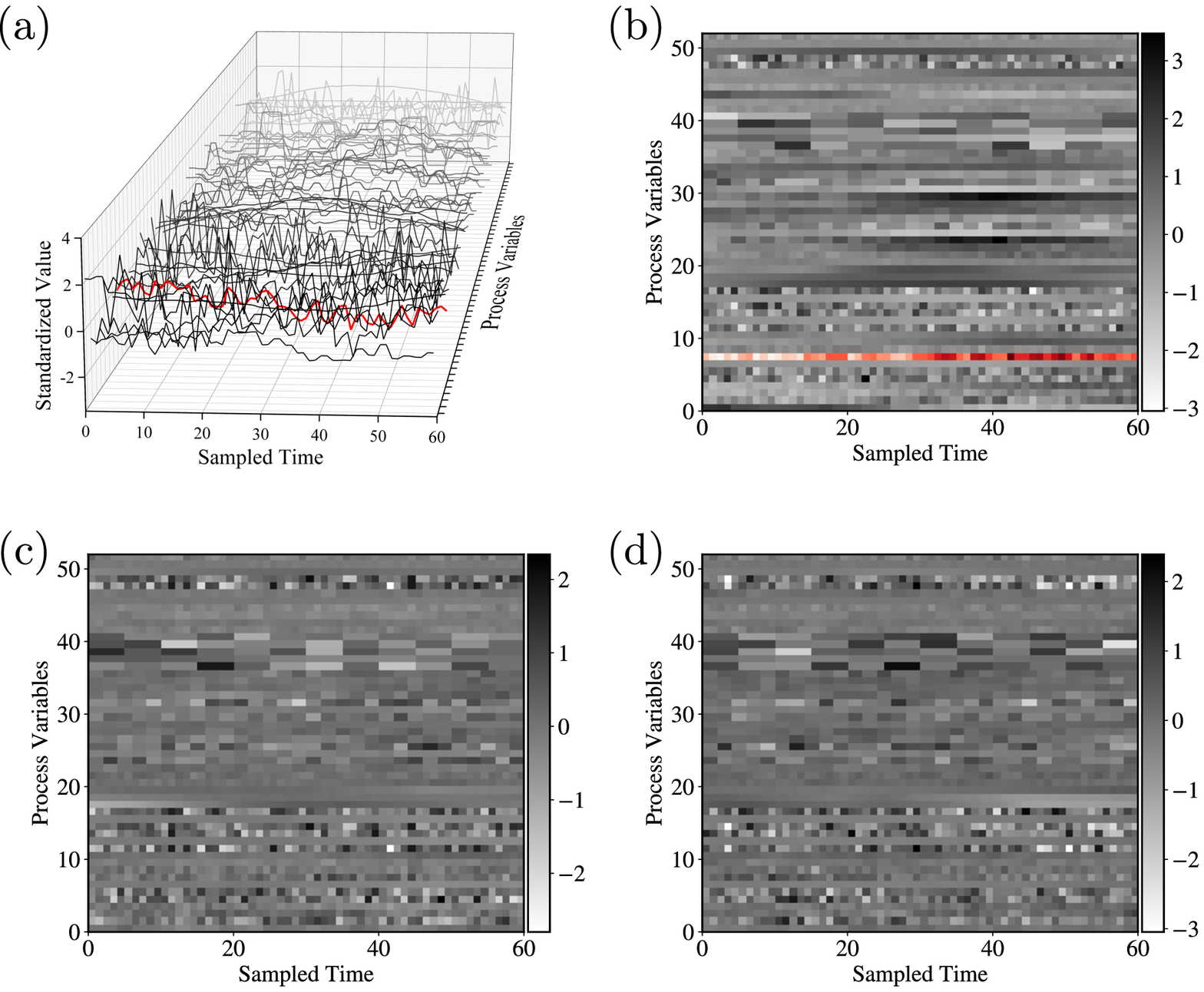}
		\caption{Transformation of multivariate signal data to 2D images. (a) 52 process variables collected 60 times over a 3-hour period. The variables are normalized with a zero mean and a unit variance. (b) An ${52 \times 60}$ matrix (visualized as an image), where the row dimension is the number of TE process variables and the column dimension is the number of time points. The red line in (a) and red row in (b) indicate the same data. Each row of the image represents one of the time series in (a). The fault number is 7 for (a) and (b), 9 for (c), and 15 for (d). We can see that (c) and (d) are visually similar but belong to different fault groups.}
		\label{fig:case2signal2heatmap}
	\end{center}
\end{figure}

The TE process data is obtained from Harvard Dataverse \cite{6C3JR1_2017}. The 52 process variables are sampled every 3 minutes.  The transformation of multivariate signal data to matrices is shown in Figure~\ref{fig:case2signal2heatmap}. In the transformation, we construct an input data sample by using 52 signal vectors (each vector contains 60 time points) that are combined into ${52 \times 60}$ matrix ($V$). We have a total of 6947 input samples. The training:validation:testing data ratio used is 11:4:5. The model architecture is Conv(64)-MaxPool-Conv(64)-MaxPool-Flatten-Dense(128)-Dense(64)-Dense(20).
\\

\begin{figure}[htb!]
	\begin{center}
		\includegraphics[width=0.6\linewidth]{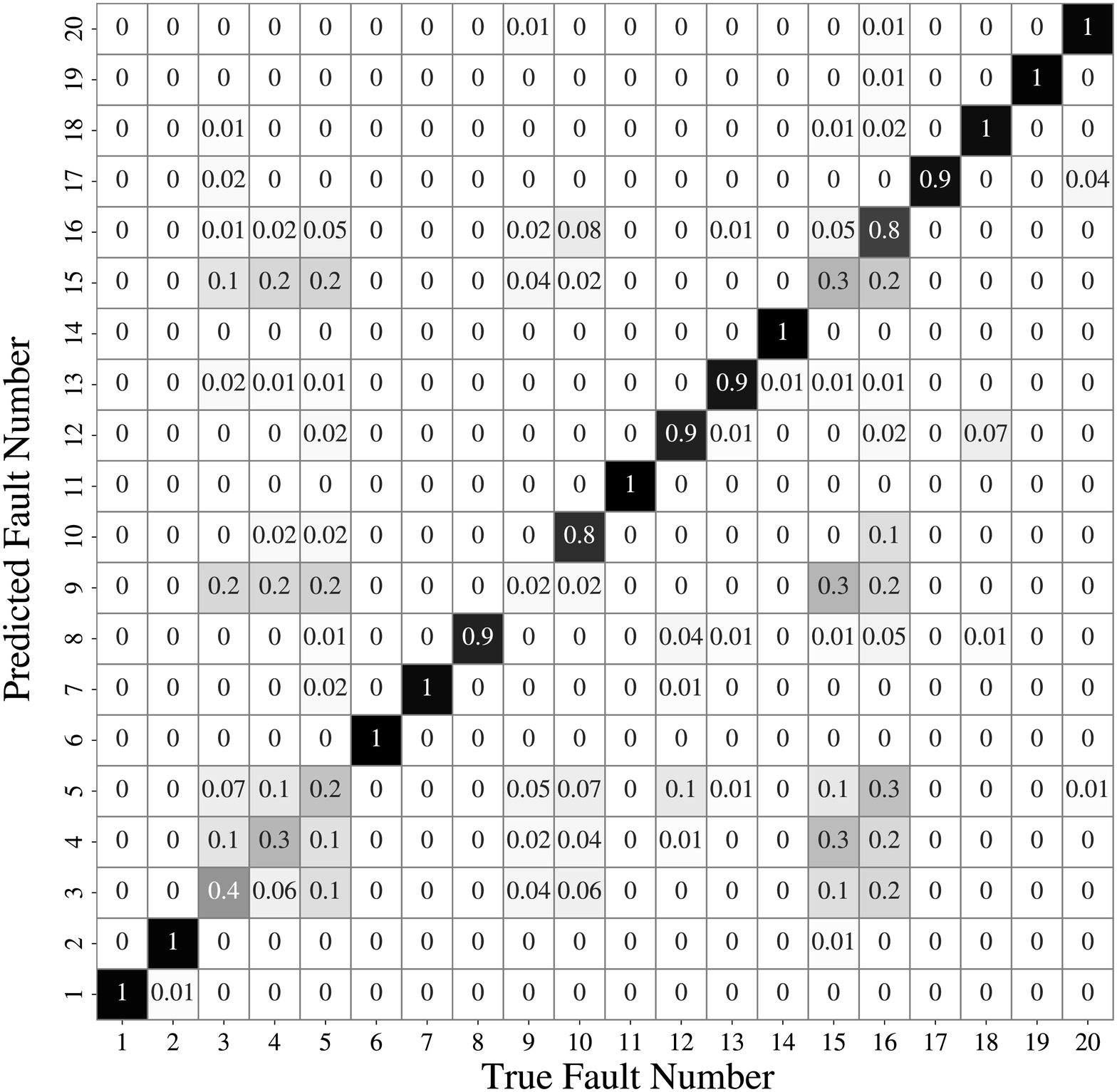}
		\caption{Confusion matrix from the CNN prediction. Each column represents a true fault type, and each row represents a CNN predicted fault type. The entries along the diagonal are where the fault types are correctly classified. Most of the diagonal entries are close to 1, indicating that the CNN has good classification accuracy.}
		\label{fig:case2confusion}
	\end{center}
\end{figure}

{\color{black} The confusion matrix is a visualization of performance in classification models. Each row of the confusion matrix represents instances of the predicted class and each column represents instances of the true class. The entries along the diagonal lines are where the instances are correctly classified.} Figure~\ref{fig:case2confusion} is a confusion matrix with an accuracy of 0.7561. With the exception of faults 3, 4, 5, 9, and 15, most faults can be identified accurately. Fault 3, 9 and 15 are especially difficult to detect because the mean, variance, and higher-order variances do not vary significantly \cite{chiang2000fault, zhang2009enhanced}.
\\

\begin{figure}[htb!]
	\begin{center}
		\includegraphics[width=1\linewidth]{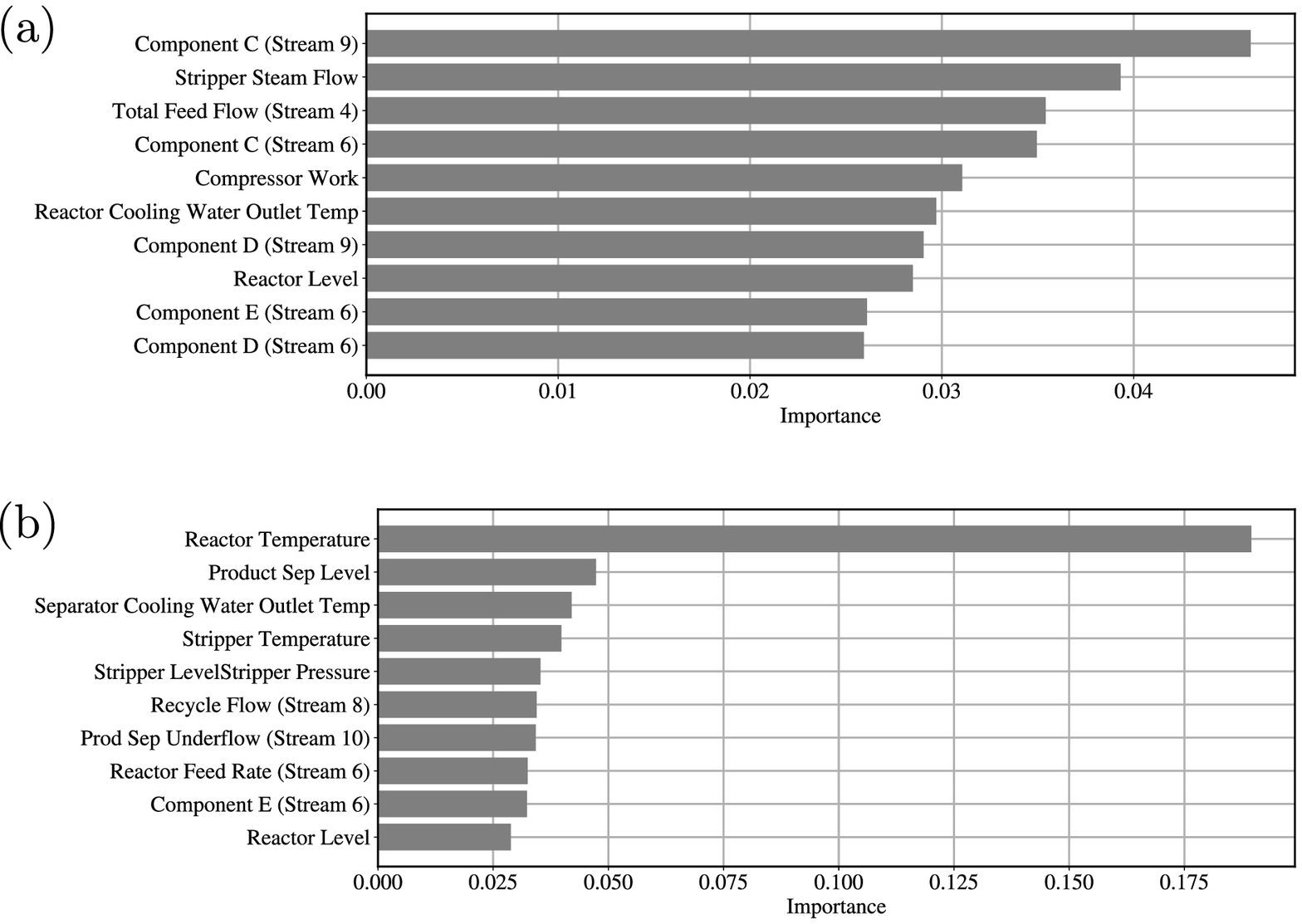}
		\caption{Input matrices (visualized as images) and their corresponding saliency maps. (a)(c) The input image with a size of $\bR^{52 \times 60}$ that belongs to fault type 7 and 8, respectively. (b)(d) The corresponding saliency maps calculated by Integrated gradients for input image in (a) and (c), respectively. (b) shows that the 9$^\mathrm{th}$ TE process variable in (a) is most important for the classification between sampled time 4 and 54. (d) shows the 43$^\mathrm{rd}$ TE process variable between sampled time 20 and 45 is most important.}
		\label{fig:case2saliency}
	\end{center}
\end{figure}

\begin{figure}[htb!]
	\begin{center}
		\includegraphics[width=1\linewidth]{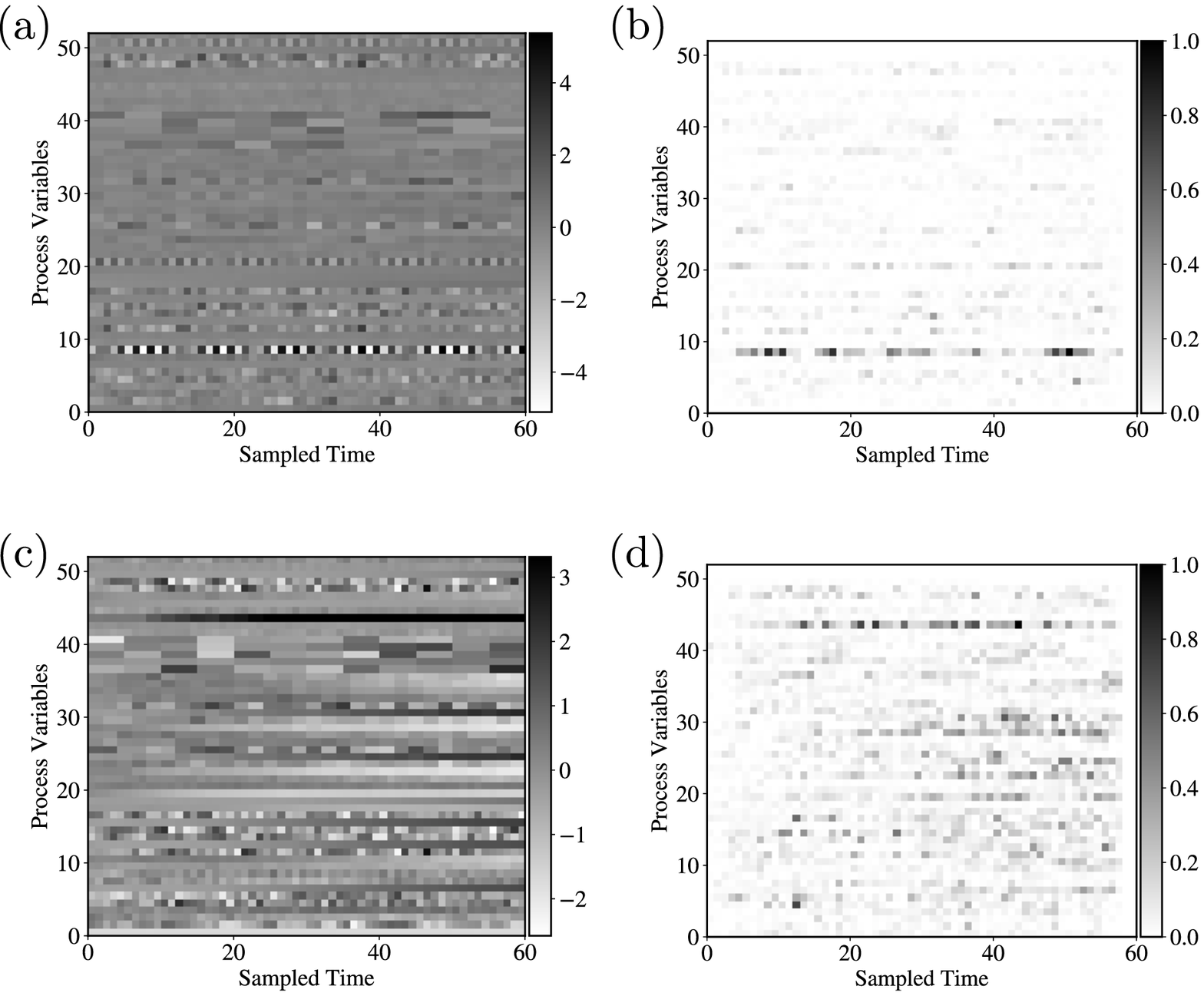}
		\caption{TE process variable saliency. (a) Most important TE process variables for detecting fault 8 (random variation of A, B, C feed composition in stream 4). The important variables are mostly related to the chemical composition in different streams. (b) Ten most important TE process variables for detecting fault 11 (random variation of reactor cooling water inlet temperature). The most important variable is the reactor temperature.}
		\label{fig:case2saliency2}
	\end{center}
\end{figure}

The saliency maps in this case study are calculated using integrated gradients; the maps are shown in Figure~\ref{fig:case2saliency}. The saliency maps reveal the most important TE process variables and time points for classification. We also performed saliency analysis for each fault. Specifically, we obtain the time-averaged saliency vector $s^i \in \bR^{52}$ from the saliency map $S^i \in \bR^{52 \times 60}$ belonging to fault type $i$. The $j$-th entry $s_j^i$ represents the importance of the $j$-th process variable in fault type $i$. We sum all the saliency vectors that are in the same fault type and rank the most important among the 52 process variables. The results are shown in Figure~\ref{fig:case2saliency2}; when the fault type is the random variation of A, B, C feed composition in stream 4, important variables are mostly related to the chemical composition in different streams (such as the compositions of component C, D, E in stream 6 and 9). This is because, in general, changes in the composition in one stream will affect the composition in another stream. When the fault type is the random variation of reactor cooling water inlet temperature, the most important variable is the reactor temperature. This is because the inlet temperature of the reactor cooling water directly affects the reactor temperature. We can thus see that saliency maps provide valuable information that facilitate interpretation of CNN predictions.
\\

\subsection{Molecular Simulations (3D)}
This case study is based on the 3D CNN reported in \cite{chew2020fast}. Our goal here is to focus on the data representation aspects of the problem in order to highlight how grid data can be used to represent 3D fields. Liquid-phase acid-catalyzed reactions can enhance biomass into high-value chemicals, but identifying the suitable solvent mixture that affects the reaction rate is challenging. In this case study, we want to show that the atomistic configuration of the reactant-solvent environment (containing reactant, water and cosolvent) generated by molecular dynamics (MD) simulations can be exploited by 3D CNNs to achieve accurate predictions of reaction rates.
\\

\begin{figure}[htb!]
	\begin{center}
		\includegraphics[width=1\linewidth]{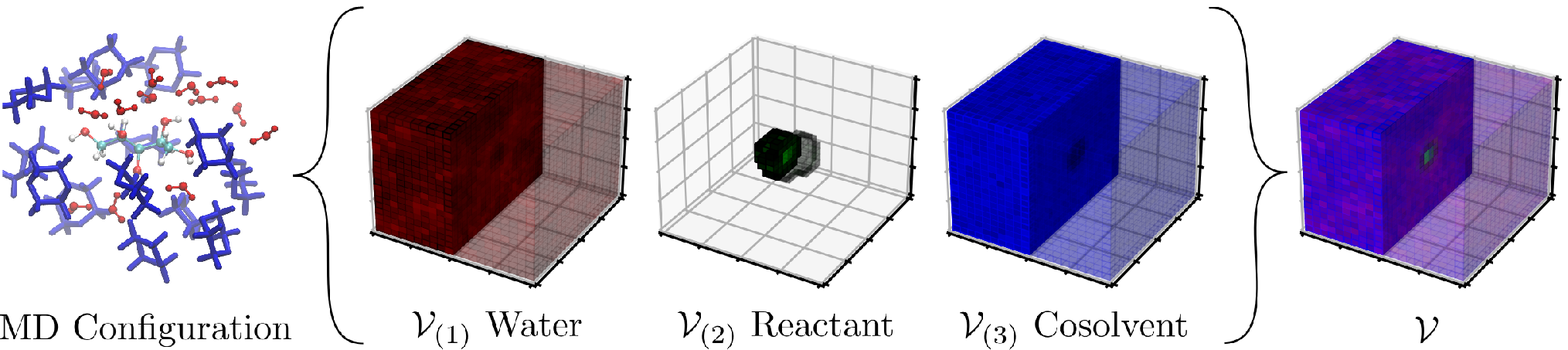}
		\caption{Approach for converting the atomic positions obtained from a MD simulation to a 3-channel tensor. For each MD configuration, a $(4 \mathrm{~nm})^3$ cubic box was centered on the reactant and a $20 \times 20 \times 20$ grid of $(0.2 \mathrm{~nm})^3$ volume elements was used to discretize space. The normalized occurrences of water, reactant, and cosolvent within each volume element were stored in different channels to yield a $20 \times 20 \times 20 \times 3$ tensor. Voxels (entries in the tensor) are visualized by showing the water channel in red, the reactant channel in green, and the cosolvent channel in blue. Half of the voxels are transparent to illustrate the solvent distribution around the reactant.}
		\label{fig:solv1}
	\end{center}
\end{figure}

Figure~\ref{fig:solv1} illustrates the data representation approach used to convert MD positions to a 3D tensor. For each MD configuration, we centered a 3D histogram on the center of mass of the reactant, which covered a cubic $(4 \mathrm{~nm})^3$ volume that was divided into a $20 \times 20 \times 20$ grid of bins (called voxels). For each bin, we calculated the water density by counting the number of water molecules within the bin and by normalizing. The same procedure was also performed to calculate the normalized occurrence of cosolvent and reactant oxygen atoms in each bin. The data object $\bV$ obtained for a single MD configuration is a 3-channel, 3D tensor. The first channel $\bV_{\rb{1}}$ is the water density field, the second channel $\bV_{\rb{2}}$ is the reactant field, and the third channel $\bV_{\rb{3}}$ is the cosolvent field. The density field of each channel is a 3D tensor of dimension $20 \times 20 \times 20$ (each entry in a tensor is a voxel); this means that the 3D tensor is a projection of a 4D probability density function (three dimensions for space and the fourth dimension is the density value). This illustrates how tensors can be used to represent high-dimensional density functions and contain large amounts of information.  We also observe that the data object $\mathcal{V}$ can be treated as a 4D tensor of dimension $20 \times 20 \times 20 \times 3$. This data representation is illustrated in Figure~\ref{fig:solv1}.  The goal here is to train a 3D CNN to identify features in the water, solvent, and cosolvent density fields that best explain reactivity.  In other words, the CNN will learn to characterize the 3D solvent environment as we vary concentrations and types of reactants, solvents, and co-solvents. In total, we obtained 18240 tensors for 76 reaction-solvent systems.
\\

\begin{figure}[htb!]
	\begin{center}
		\includegraphics[width=1\linewidth]{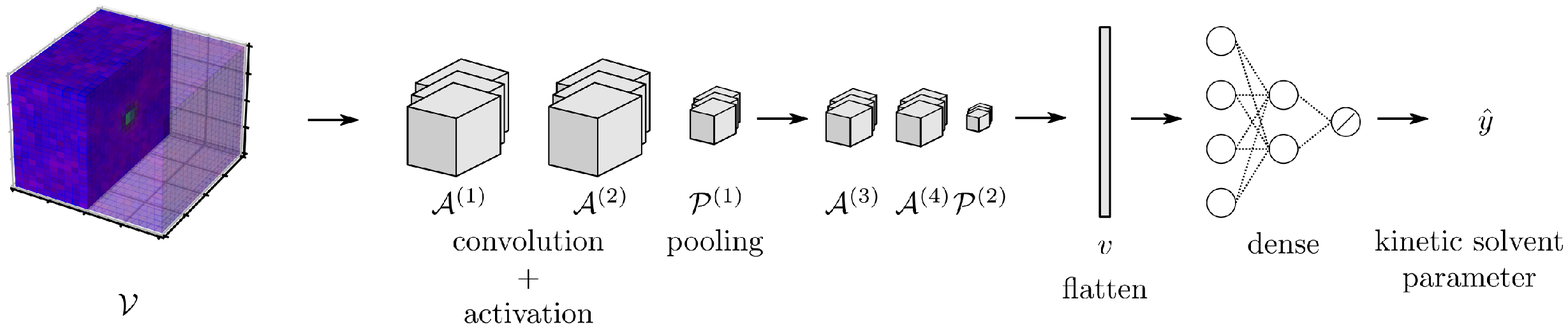}
		\caption{SolventNet architecture. The input to SolventNet is a 3-channel object $\mathcal{V} \in \bR^{20 \times 20 \times 20 \times 3}$; the channels correspond to the water, reactant and cosolvent normalized occurrences (each is a tensor of dimension $20 \times 20 \times 20$). SolventNet contains four convolutional blocks with 8, 16, 32, and 64 $3 \times 3 \times 3$ operators, respectively, and two max-pooling blocks with $2 \times 2 \times 2$ operators. The feature maps $\mathcal{A}^{(1)}$, $\mathcal{A}^{(2)}$, $\mathcal{A}^{(3)}$ and $\mathcal{A}^{(4)}$ are tensors of $\bR^{18 \times 18 \times 18 \times 8}$, $\bR^{16 \times 16 \times 16 \times 16}$, $\bR^{6 \times 6 \times 6 \times 32}$ and $\bR^{4 \times 4 \times 4 \times 64}$, respectively. The feature maps $\mathcal{P}^{(1)}$ and $\mathcal{P}^{(2)}$ are tensors of $\bR^{8 \times 8 \times 8 \times 16}$, $\bR^{2 \times 2 \times 2 \times 54}$, respectively. Later $\mathcal{P}^{(2)}$ is flattened into a vector $v \in \bR^{512}$. This vector is passed to three dense layers (each having 128 hidden units). The predicted kinetic solvent parameter $\hat{y} \in \bR$ is the output from the dense layer activated by a linear function.}
		\label{fig:solv2}
	\end{center}
\end{figure}

\begin{figure}[htb!]
	\begin{center}
		\includegraphics[width=0.6\linewidth]{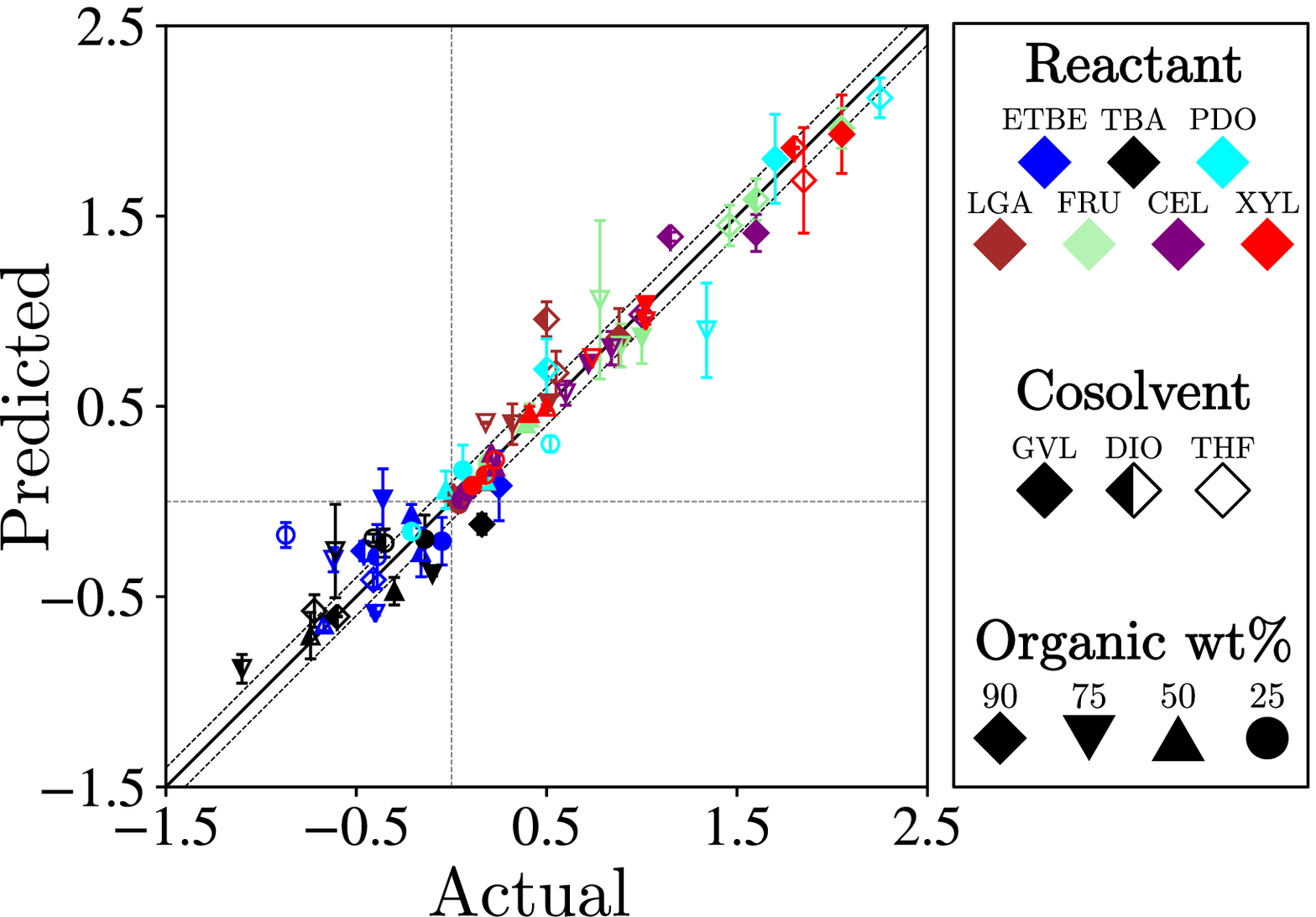}
		\caption{Predicted and actual kinetic solvent parameters using SolventNet for various reactants, cosolvents and organic weight percents. Each point is the average prediction of 10 samples per label. Error bars show the standard deviation of these predictions. The RMSE between actual and predicted values is 0.37. The solid black lines indicate perfect correlation and the dashed black lines indicate approximate experimental error. Reproduced from \cite{chew2020fast} with permission from the Royal Society of Chemistry.}
		\label{fig:solv3}
	\end{center}
\end{figure}

The 3-channel data object was fed to a 3D CNN, which we call {\tt  SolventNet}. The output block generates a scalar prediction $\hat{y}$ (corresponding to the kinetic solvent parameter). The architecture of SolventNet is shown in Figure~\ref{fig:solv2}. Regression results for SolventNet are presented in Figure \ref{fig:solv3}. The RMSE between actual and predicted values is 0.37, which is much better than the result (0.58) of previously reported methods. Specifically, SolventNet has good generality for various reactants, cosolvents and organic weight fractions, while the descriptor-based approach is ineffective in predicting systems with a large organic weight fraction.

\begin{figure}[htb!]
	\begin{center}
		\includegraphics[width=0.8\linewidth]{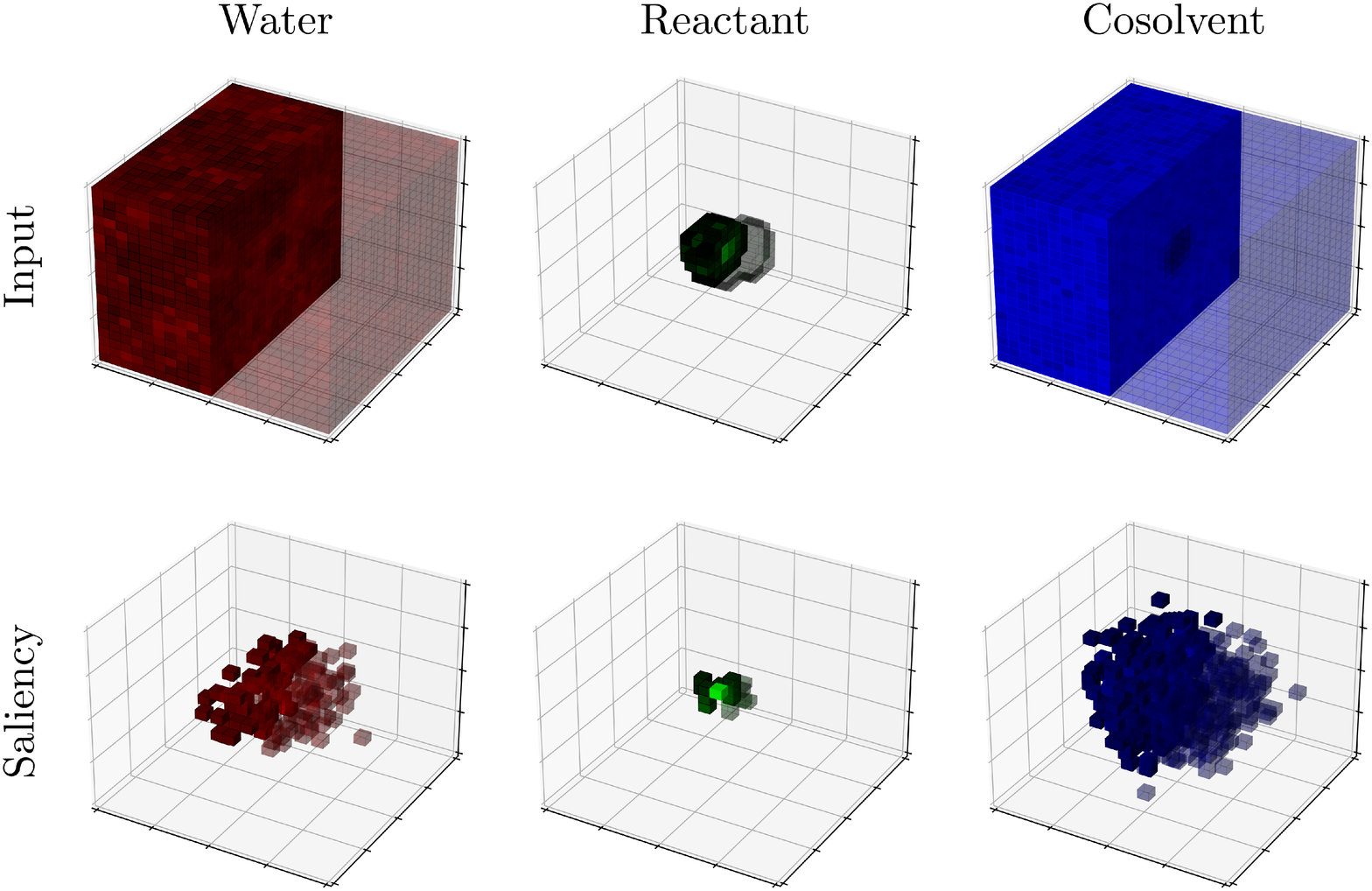}
		\caption{Saliency map from SolventNet. The saliency map (bottom) is visualized on a 3D grid with the same dimensions as the input tensor (top). Each voxel is assigned a saliency value normalized from 0 to 1 that indicates the sensitivity of SolventNet predictions to the normalized occurrences of water, reactant, and cosolvent in that voxel. The saliency maps show that the region near the reactants is the most important for prediction.}
		\label{fig:solv4}
	\end{center}
\end{figure}

The saliency map of SolventNet is calculated by integrated gradients. Examination of the saliency map in Figure~\ref{fig:solv4} confirms that SolventNet primarily identifies features of the solvent environment within the local domain near the reactant. These plots clearly show that the region near the reactants is the most important for prediction, and that the size of the simulated volume is large enough that the region far from the reactants is not important.
\\

\section{Conclusions and Future Work}\label{sec:conclusion}

Convolutional neural networks (CNNs) are machine learning models that analyze grid-like data to unravel hidden patterns and features to make predictions. We have shown that data representations used by CNNs offer significant flexibility to tackle different types of applications (that go beyond computer vision). We have provided a review of key concepts behind CNNs (convolutions, multi-channel signals, operators, forward/backward propagations, and saliency maps) under a unifying mathematical framework. Our objective here is to establish connections with concepts from other mathematical fields and to highlight key computations that take place inside these powerful models. 
\\

There are several opportunities to extend CNNs from an algorithmic stand-point. We have seen that the parameters of CNNs are tensors and thus a huge number of parameters need to be optimized; as such, strategies for regularization can be envisioned. Specifically, one can envision optimizing operators with predefined  structures.  The connections between differential operators and convolutions are also interesting; here, one can envision developing architectures with fixed differential operators that are combined to learn structures of partial differential equations. Opportunities also exist for using CNNs in new applications. For instance, the ability of 1D CNNs to analyze sequences could be used to build dynamic models; currently, such models are being developed using fully-connected neural nets (but CNNs are more effective at capturing time dependencies). There is also potential in using CNNs to analyze space-time datasets; as those arising in computational fluid dynamics and other PDE applications. In this context, a key computational challenge is generalizing CNNs to 4D (while capturing  high resolutions). One possibility could be the creation of hybrid CNN models that partition the input domain into subdomains of high resolution and that combine this with information of the full domain at low resolution. 
\\

{\color{black} While CNNs are capable of analyzing grid-like data, data in science and engineering can also appear in non-grid form (e.g. graphs). For example, bonds in molecules can be interpreted as edges in a graph, while atoms are nodes. Recently, graph neural networks (GNNs) have become very popular for analyzing graph-structured data. GNNs have been successfully used to predict molecular quantum properties \cite{gilmer2017neural}, reactivity \cite{coley2019graph} and other chemical properties \cite{yang2019analyzing}. However, the full application potential of GNNs in chemical engineering has not been explored. In addition, it would be useful to establish fundamental connections between CNNs and GNNs; specifically, CNNs are special classes of GNNs. Understanding such connections can help unify these powerful modeling paradigms.}


\section*{Acknowledgments}
We acknowledge funding from the U.S. National Science Foundation (NSF) under BIGDATA grant
IIS-1837812 and also acknowledge partial support from the NSF through the University of Wisconsin
Materials Research Science and Engineering Center (DMR-1720415). We thank Prof. Nicholas Abbott for providing the endotoxin detection dataset and Prof. Reid Van Lehn for providing the molecular simulation dataset. 

\clearpage
\newpage
\bibliography{CNN}

\end{document}